\documentclass{article}

\usepackage[a4paper, total={6in, 8in}]{geometry}
\usepackage[utf8]{inputenc}
\usepackage{authblk}
\usepackage{color}
\usepackage{enumitem}

\usepackage{amsfonts}
\usepackage{amsmath}
\usepackage{amssymb}
\usepackage{amsthm}
\usepackage{bm}
\usepackage{mathtools}

\usepackage{graphicx}
\usepackage{epstopdf}
\usepackage{subcaption}

\definecolor{grey}{rgb}{.7,.7,.7}

\definecolor{evidGP}{rgb}{0,0,1}  

\definecolor{evidM}{rgb}{0,0.5,0}  

\theoremstyle{definition}
\newtheorem{dfn}{Definition}
\newtheorem{xmp}{Example}

\theoremstyle{plain}
\newtheorem{thm}{Theorem}[section]

\newtheorem{lem}[thm]{Lemma}
\newtheorem{cor}[thm]{Corollary}

\theoremstyle{remark}
\newtheorem{rmk}{Remark}

\newcommand{\E}{\mathbb{E}}             
\newcommand{\N}{\mathbb{N}}             
\newcommand{\R}{\mathbb{R}}             
\renewcommand{\l}{\ell}                 
\newcommand{\ph}{\varphi}               
\newcommand{\x}{\mathbf{x}}             
\newcommand{\m}{\mathbf{m}}             
\renewcommand{\hm}{\mathbf{\hat{m}}}    
\newcommand{\hM}{\widehat{M}}           
\newcommand{\xx}{\bm{\xi}}              
\newcommand{\w}{\mathbf{w}}             
\newcommand{\W}{\mathbf{W}}             
\renewcommand{\b}{\mathbf{b}}           
\newcommand{\s}{\mathbf{s}}             
\newcommand{\Loss}{\mathcal{L}}         
\newcommand{\eps}{\varepsilon}
\newcommand{\ds}{\displaystyle}

\DeclareMathOperator{\argmax}{\arg\max}
\DeclareMathOperator{\Lip}{Lip}
\DeclareMathOperator{\diam}{diam}

\graphicspath{{./}}

\title{\bf{Analytical aspects \\ of non-differentiable neural networks}}
\date{}
\author[1]{Gian Paolo Leonardi}
\author[2]{Matteo Spallanzani}
\affil[1]{\footnotesize Dipartimento di Matematica, Universit\`{a} di Trento. Via Sommarive 14, 38123 Trento, Italy}
\affil[2]{\footnotesize Departement Informationstechnologie und Elektrotechnik, ETH Z\"{u}rich. Gloriastrasse 35, 8092 Z\"{u}rich, Switzerland}

\begin{document}

\maketitle

\begin{abstract}
\noindent
Research in computational deep learning has directed considerable efforts towards hardware-oriented optimisations for deep neural networks, via the simplification of the activation functions, or the quantization of both activations and weights.
The resulting non-differentiability (or even discontinuity) of the networks poses some challenging problems, especially in connection with the learning process.
In this paper, we address several questions regarding both the expressivity of quantized neural networks and approximation techniques for non-differentiable networks.
First, we answer in the affirmative the question of whether QNNs have the same expressivity as DNNs in terms of approximation of Lipschitz functions in the $L^{\infty}$ norm.
Then, considering a continuous but not necessarily differentiable network, we describe a layer-wise stochastic regularisation technique to produce differentiable approximations, and we show how this approach to regularisation provides elegant quantitative estimates.
Finally, we consider networks defined by means of Heaviside-type activation functions, and prove for them a pointwise approximation result by means of smooth networks under suitable assumptions on the regularised activations.
\end{abstract}
\bigskip


\section{Introduction}
\label{sec:intro}

Deep neural networks (DNNs) have established a prominent position amongst machine learning (ML) systems \cite{Hinton2007, Bengio2009}.
Most of this success has been due to the fact that DNNs have enabled super-human performance of artificial intelligence (AI) in multiple complex domains, including computer vision \cite{Krizhevsky2012, He2016}, natural language processing \cite{Vaswani2017, Devlin2019}, and decision making in both complete and incomplete information games \cite{Silver2017, Vinyals2019}.
Other critical factors of this success have been the computational properties enjoyed by DNNs.
Seen as computer programs, DNNs have data-independent control flows (i.e., they avoid branching instructions) and are highly parallelisable, both at the neuron level and at the layer level.  
These properties are desirable since they enable efficient memory access patterns (the so-called \textit{spatial} and \textit{temporal locality}), they can take advantage of vector instructions available on modern \textit{single instruction, multiple data} (SIMD) computer architectures, and they expose a level of concurrency suitable even for distributed computing systems.

To achieve their outstanding performance, most DNNs need millions or billions of trainable parameters and must perform a proportional amount of arithmetic operations.
These characteristics make DNNs very demanding programs in terms of storage, memory, and energy, both during training and inference, most times preventing their deployment on low-memory and low-power computers such as embedded devices and microcontroller units (MCUs).
Thus, alleviating the pressure of DNNs on the computing infrastructure while preserving accuracy is an important research topic in computational deep learning.
The solutions to this problem can be categorised into two main families:
\begin{itemize}
    \item \textit{topological optimisations}, i.e., designing network topologies which are more efficient in terms of accuracy-per-parameter or accuracy-per-MAC (multiply-accumulate operation) \cite{Howard2017, Sandler2018, Ma2018, Zhang2018};
    \item \textit{hardware-oriented optimisations}, i.e., choosing more hardware-efficient alternatives for the activation functions and more compact numerical representations for the parameters \cite{Nair2010, Hubara2018}.
\end{itemize}

Hardware-oriented optimisations aim at taking advantage of properties that are specific to digital hardware.
For example, the \textit{instruction set architectures} (ISAs) of many RISC processors do not include specialised instructions to evaluate transcendent functions.
Such evaluations are usually performed by low-level numerical libraries, which result in a much higher latency than evaluating less regular functions like the piece-wise affine rectified linear unit (ReLU) or the piece-wise constant Heaviside.
In fact, replacing sigmoid, differentiable activation functions with ReLUs has drastically reduced the training time for DNNs, while at the same time preserving or even increasing statistical accuracy \cite{Krizhevsky2012}.
It is therefore clear how hardware-oriented optimisations have been playing an important role in favouring the widespread adoption of DNNs.

More recently, the need to deploy DNN-based applications onto embedded devices and MCUs has motivated the investigation of \textit{quantized neural networks} (QNNs).
QNNs are specific DNNs whose parameters take values in small, finite sets and whose activation functions have a finite range.
These features allow representing the operands with fewer bits with respect to standard DNNs, and this can lead to multiple benefits when considering digital hardware.
First, given a budget of storage and memory, it is possible to fit more parameters and activations of a QNN than it is possible for a DNN with the same topology.
Second, using fixed-point representations for the operands allows replacing floating-point arithmetic with the more efficient integer arithmetic or even specialised instructions.
Third, when the operands of a dot product operation take values in the set $\{-1, 0, 1\}$, it is possible to avoid multiplications, resulting in considerable energy savings.

However, hardware-oriented optimisations also bring along theoretical subtleties which have not yet been thoroughly investigated.
In particular, questions arise about the expressivity of these networks and their training algorithms.

Classical results on the expressivity properties of DNNs assume networks with continuous-valued parameters \cite{Cybenko1989, Hornik1991}.
Despite the vast literature on experimental research about QNNs, we were not able to find any theoretical result supporting the belief that QNNs can be a valid alternative to DNNs.

Standard DNNs are built under the assumption that the layer maps are differentiable, an assumption that is typically violated by networks adopting hardware-related optimisations.
The standard supervised learning algorithm for DNNs uses the gradients as learning signals, and some variant of gradient descent \cite{Nocedal2006} as the learning rule to update the parameters.
Under the hypothesis that each layer map is differentiable, the compositional structure of DNNs implies that the gradients can be computed using the chain rule.
This feature makes DNNs suitable candidates for the application of \textit{automatic differentiation} methods such as the backpropagation algorithm \cite{Rumelhart1986}.
The most popular software frameworks for DNN training such as TensorFlow \cite{Abadi2016} and Pytorch \cite{Paszke2017} have therefore been designed according to the \textit{dataflow} programming model \cite{Dennis1974} to take advantage of these structural properties.
In this programming model, a network is represented as a \textit{computational graph}, a directed graph whose nodes represent functions and whose edges represent dependencies between functions.
When each layer map is supposed to be differentiable, each node in the graph represents a differentiable function.
The advantage of these \textit{differentiable computational graphs} is that gradient computation by \textit{reverse-mode automatic differentiation} (i.e., backpropagation) is algorithmically similar to the inference (i.e., forward) computation.
This formal analogy allows for reusable library functions, simpler code maintenance, and efficient mapping to diverse computing infrastructures, making the frameworks built on top of dataflow abstractions extremely efficient.
On the one hand, developers would benefit from the speed of these optimised frameworks, for rapid prototyping and application development.
On the other hand, this constraint calls for correct strategies to solve non-differentiable or even discontinuous optimisation problems using differentiable computational graphs.
Therefore, in this work, we focus on \textit{layer-wise regularisations} of non-differentiable DNNs.

In the community of machine learning practitioners, the experimental success of ReLU-activated networks has contributed to developing confidence in the fact that isolated non-differentiabilities are not essential issues for gradient-based learning algorithms.
But, to the best of our knowledge, we do not know of any use of generalised notions of gradient (like, e.g., the sub-gradient, see \cite{Shor1985}) in optimising compositions of non-differentiable functions.

More recently, research in QNNs has applied gradient-based training algorithms to networks including discontinuous operations such as the Heaviside function \cite{Hubara2018}.
In principle, any DNN topology using discontinuous activations of Heaviside-type (hence, activations whose derivative vanishes almost-everywhere) cannot be directly trained via classical or stochastic gradient-based methods.
In this case, empirical results have shown variable effectiveness, suggesting that the transition from localised non-differentiabilities to discontinuities can indeed introduce significant errors.
Interestingly, this problem is related to some of the problems that arose in seminal research on artificial neural networks.
McCulloch and Pitts analyzed networks of neurons using discrete weights and binary activation functions \cite{McCulloch1943}; Rosenblatt used both quantized weights and quantized activations for his \textit{perceptron} \cite{Rosenblatt1957}; Widrow and Hoff used continuous-valued parameters but binary activations for their \textit{ADALINE} \cite{Widrow1960}.
When the search for a suitable rule to train multi-layer networks composed of such units clashed with the difficulty of deriving proper error-correcting signals, the adoption of differentiable activation functions broke the stall, opening the way to the application of gradient-based training of multi-layer networks \cite{Werbos1974, Rumelhart1986}.

\bigskip
The paper is organised as follows.
\begin{itemize}
    \item Section~\ref{sec:notation} contains the terminology and the notations used throughout the paper. 
    \item In Section~\ref{sec:expressivity} we prove Theorem \ref{th:expressivity}, a universal approximation result for QNNs, showing that quantized networks enjoy the same expressivity of DNNs using continuous parameters and Lipschitz activation functions.
        We also provide the explicit estimate \eqref{eq:model_size_bound} on the number of neurons that are necessary to achieve a desired degree of approximation.
    \item In Section~\ref{sec:non-differentiability} we show that the problem of training a DNN defined by means of non-differentiable activation functions can be approximated by a sequence of training problems for regularised (hence, differentiable) DNNs.
        More specifically, Theorem~\ref{th:uniform_compositional_approximation} provides a quantitative approximation of continuous (but not necessarily differentiable) networks by means of smooth networks obtained via layer-wise stochastic regularisation (see Section \ref{subsub:stochastic} for the essential definitions).
        In the special case of Lipschitz networks, one deduces a \textit{variance annealing schedule} for equalised approximation, that is, a suitable choice of the variances of the stochastic layer maps ensuring that the composition of the first $\l$ layer maps is approximated up to a uniform error bound, for each $\l = 1, \dots, L$ (see Remark~\ref{rmk:uniform_compositional_approximation_annealing}).
    \item In Section~\ref{sec:discontinuity} we provide sufficient conditions for approximating (in pointwise sense) a QNN defined by means of Heaviside-type activation functions by a sequence of networks that use suitably regularised activation functions.
        The key notion that is introduced in this section is the \textit{rate convergence} property (see Definition~\ref{dfn:rate_convergence}).
        In Theorem \ref{th:pointwise_convergence} we show that a layer-wise regularisation of a DNN using the Heaviside function $H^{+}$ as its activation function pointwise converges to the original network as soon as the regularised activations satisfy the rate convergence assumption.
        In this respect, we also provide Examples \ref{xmp:pointwise_convergence_xmp} and \ref{xmp:pointwise_convergence_counterxmp} showing the tightness of the above definition and result.
        We then conclude the section with Example \ref{xmp:continuous_to_quantized_parameters}, where we discuss a potential problem with current training algorithms for QNNs.
\end{itemize}

\section{Notation and definitions}
\label{sec:notation}

\subsection{Feedforward neural networks}
Let $L \geq 2$ be an integer representing the \textbf{number of layers} of the network.
For every $\l = 0, 1, \dots, L$, we define non-empty sets $X^{\l}$ and $M^{\l}$, called the $\l$-th \textbf{representations space} and the $\l$-th \textbf{parameters space}, respectively.
For every $\l = 1, \dots, L$, the function
\begin{equation*}
\begin{split}
    \ph^{\l} \,:\,
    X^{\l-1} \times M^{\l} &\to X^{\l} \\
    (\x^{\l-1}, \m^{\l}) &\mapsto \x^{\l} \coloneqq \ph^{\l}(\x^{\l-1}, \m^{\l})
\end{split}
\end{equation*}
is called the $\l$-th \textbf{layer map}.
To simplify the notation when $\m^{\l} \in M^{\l}$ is fixed, we define
\begin{equation}\label{eq:layer_map}
\begin{split}
    \ph_{\m^{\l}} \,:\,
    X^{\l-1} &\to X^{\l} \\
    \x^{\l-1} &\mapsto \ph_{\m^{\l}}(\x^{\l-1}) \coloneqq \ph^{\l}(\x^{\l-1}, \m^{\l}) \,.
\end{split}
\end{equation}

Given $\l \in \{ 1, \dots, L \}$, we define $\hm^{\l} \coloneqq (\m^{1}, \m^{2}, \dots, \m^{\l})$ to be the collective parameter taken from the space $\hM^{\l} \coloneqq M^{1} \times M^{2} \times \dots \times M^{\l}$. 
This allows us to give the recursive definition
\begin{align}
    \Phi_{\hm^{1}} &\coloneqq \ph_{\m^{1}} \,, \nonumber \\
    \Phi_{\hm^{\l}} &\coloneqq \ph_{\m^{\l}} \circ \Phi_{\hm^{\l-1}} \,,\, \l = 2, \dots, L \,, \label{eq:composition_of_layer_maps}
\end{align}
where $\circ$ denotes composition.
The complete feedforward network is denoted by
\begin{equation}\label{eq:deep_neural_network}
    \Phi \coloneqq \Phi_{\hm^{L}} \,.
\end{equation}

The notation defined so far is general enough to represent any compositional chain of parametric maps.
However, in practice, the representations spaces, the parameters spaces, and the layer maps are more specific.
For every $\l = 0, 1, \dots, L$ we denote by the positive integer $n_{\l} \in \N$ the \textbf{number of neurons} in the $\l$-th layer.
Then, the $\l$-th representations space is modelled as a subset of the $n_{\l}$-dimensional Euclidean space: $X^{\l} \subseteq \R^{n_{\l}}$.
The $\l$-th parameters space are instead set products $M^{\l} \coloneqq W^{\l} \times B^{\l}$, where
\begin{align}
    W^{\l} &\subseteq \R^{n_{\l-1} \times n_{\l}} \,, \label{eq:weights_space} \\
    B^{\l} &\subseteq \R^{1 \times n_{\l}} \,, \label{eq:biases_space}
\end{align}
are called the $\l$-th \textbf{weights space} and the $\l$-th \textbf{bias space}, respectively; with this notation, $\m^{\l} \coloneqq (\W^{\l}, \b^{\l})$.

We then define the affine map
\begin{equation*}
\begin{split}
    A^{\l} \,:\,
    X^{\l-1} \times M^{\l} &\to \R^{n_{\l}} \\
    (\x^{\l-1}, \m^{\l}) &\mapsto \s^{\l} \coloneqq \x^{\l-1} \W^{\l} + \b^{\l} \,,
\end{split}
\end{equation*}
where $\x^{\l-1} \W^{\l}$ is a vector-matrix product.
Analogously to \eqref{eq:layer_map}, given $\m^{\l} \in M^{\l}$, we define
\begin{equation}\label{eq:layer_map_affine_part}
\begin{split}
    A_{\m^{\l}} \,:\,
    X^{\l-1} &\to \R^{n_{\l}} \\
    \x^{\l-1} &\mapsto A_{\m^{\l}}(\x^{\l-1}) \coloneqq \s^{\l} \,.
\end{split}
\end{equation}

Let
\begin{equation}\label{eq:activation_function}
\begin{split}
    \sigma \,:\,
    \R &\to \R \\
    s &\mapsto \sigma(s)
\end{split}
\end{equation}
denote a real function that we name the \textbf{activation function}, which should be non-constant and non-decreasing; if it is also bounded, we say that it is a \textit{sigmoid} activation function.
Typical examples of sigmoid activation functions are the \textit{logistic}
\begin{equation*}
    \sigma(s) \coloneqq \frac{1}{1 + e^{-s}} \,,
\end{equation*}
and the \textit{hyperbolic tangent}
\begin{equation*}
    \sigma(s) \coloneqq \frac{e^{s} - 1}{e^{s} + 1} \,.
\end{equation*}
Other common activation functions are the \textit{rectified linear unit} (ReLU)
\begin{equation*}
    \sigma(s) \coloneqq
    \begin{cases}
        0 \,, &\text{if } s < 0 \,, \\
        s \,, &\text{if } s \geq 0 \,,
    \end{cases}
\end{equation*}
the Heaviside functions
\begin{align}
    H^{+}(s) &\coloneqq
    \begin{cases}
        0 \,, &\text{if } s < 0 \,, \\
        1 \,, &\text{if } s \geq 0 \,,
    \end{cases} \label{eq:heaviside_zero2one} \\
    H^{-}(s) &\coloneqq
    \begin{cases}
        0 \,, &\text{if } s \leq 0 \,, \\
        1 \,, &\text{if } s > 0 \,,
    \end{cases} \label{eq:heaviside_zero2zero}
\end{align}
and even the identity (also known as the \textit{linear activation function})
\begin{equation*}
    \sigma(s) \coloneqq s \,.
\end{equation*}
\begin{figure}
    \centering
    \includegraphics[width=0.7\textwidth]{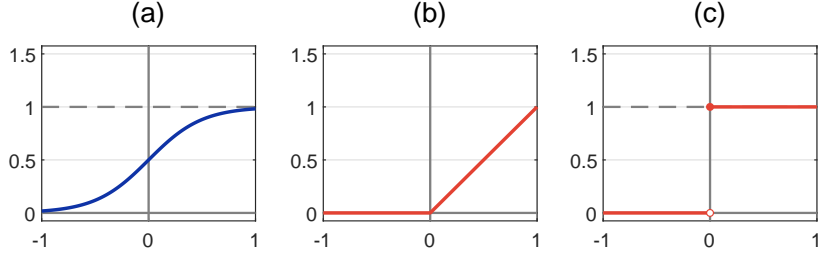}
    \caption{Examples of activation functions: (a) the logistic is smooth, (b) the piecewise linear ReLU has a non-differentiability in $s = 0$, (c) the piecewise constant Heaviside $H^{+}$ has a discontinuity in $s = 0$.}\label{fig:activation_functions}
\end{figure}
%
Let $n$ be a positive integer, and let $\sigma_{1}, \sigma_{2}, \dots, \sigma_{n}$ denote activation functions.
We define the $n$-dimensional activation map as the vector function
\begin{equation}\label{eq:layer_map_non-linear_part}
\begin{split}
    \bm{\sigma} \,:\
    \R^{n} &\to \R^{n} \\
    \s &\mapsto \left( \sigma_{1}(s_{1}), \sigma_{2}(s_{2}), \dots, \sigma_{n}(s_{n}) \right) \,,
\end{split}
\end{equation}
which applies the $i$-th activation function to the $i$-th component of the $n$-dimensional Euclidean vector $\s$. 

With these specifications, the typical layer map is built as the composition of an affine map $A_{\m^\l}$ and a non-linear activation $\bm{\sigma}$:
\begin{equation}\label{eq:layer_map_decomposition}
\begin{split}
    \ph_{\m^{\l}} \,:\,
    X^{\l-1} &\to X^{\l} \\
    \x^{\l-1} &\mapsto \x^{\l} \coloneqq (\bm{\sigma} \circ A_{\m^{\l}})(\x^{\l-1}) \,.
\end{split}
\end{equation}
We observe, though, that in both applications and theoretical research the last layer map is usually supposed to be affine or even just linear:
\begin{equation*}
    \ph_{\m^{L}} = A_{\m^{L}}
\end{equation*}
(i.e., the non-linearity \eqref{eq:layer_map_non-linear_part} is not applied).

\subsection{Quantized neural networks}
\label{sub:qnns}
Let $K \geq 2$ be an integer.
The finite set of real numbers
\begin{equation}\label{eq:quantization_set}
    Q \coloneqq \{ q_{1} < q_{2} < \dots < q_{K} \} \subset \R
\end{equation}
is called a \textbf{$K$-quantization set}.
The elements $q_{k}, k = 1, 2, \dots, K$ are called \textbf{quantization levels}.

We say that a layer map \eqref{eq:layer_map_decomposition} is \textbf{$Q$-quantized by weights} if its affine part \eqref{eq:layer_map_affine_part} is such that
\begin{equation*}
    W^{\l} = Q^{n_{\l-1} \times n_{\l}} \,;
\end{equation*}
i.e., if its weights take values in $Q$.

We say that an activation function \eqref{eq:activation_function} is a \textbf{$Q$-quantizer} if its codomain is $Q$:
\begin{equation*}
    \sigma \,:\, \R \to Q \,.
\end{equation*}
\begin{xmp}
If we set $Q = \{ 0, 1 \}$, the Heaviside functions \eqref{eq:heaviside_zero2one}, \eqref{eq:heaviside_zero2zero} are $\{ 0, 1 \}$-quantizers.
\end{xmp}
\begin{xmp}
Given a $K$-quantization set $Q = \{ q_{1}, \dots, q_{K} \}$, define the \textit{jumps} between the quantization levels as
\begin{equation*}
    d_{k-1}^{k} \coloneqq q_{k} - q_{k-1} \,,\, k = 2, 3, \dots, K \,.
\end{equation*}
Let
\begin{equation*}
    \Theta \coloneqq \{ \theta_{2} < \theta_{3} < \dots < \theta_{K} \} \subset \R
\end{equation*}
denote a set of real \textit{thresholds}.
If we define
\begin{equation}
    H_{\theta}(s) \coloneqq H^{+}(s - \theta) \,,
\end{equation}
the \textit{stair function}
\begin{equation}
    \sigma(s) \coloneqq q_{1} + \sum_{k=2}^{K} d_{k-1}^{k} H_{\theta_{k}}(s)
\end{equation}
is a $\{ q_{1}, \dots, q_{K} \}$-quantizer.
An instance of stair function is depicted in Figure~\ref{fig:stair_function}.
\begin{figure}
    \centering
    \includegraphics[height=7cm]{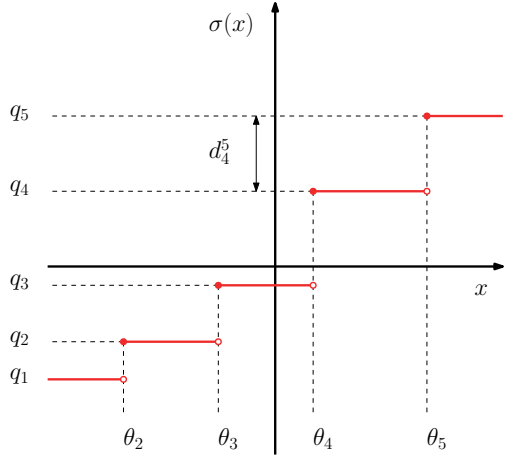}
    \caption{The stair function is an example of $Q$-quantizer.}\label{fig:stair_function}
\end{figure}
\end{xmp}
We say that a layer map \eqref{eq:layer_map_decomposition} is \textbf{$Q$-quantized by activations} if its non-linear part \eqref{eq:layer_map_non-linear_part} is the element-wise application of $Q$-quantizers.
In this way, the corresponding representations space takes the form
\begin{equation*}
    X^{\l} = Q^{n_{\l}} \,.
\end{equation*}

We say that a layer map \eqref{eq:layer_map_decomposition} is \textbf{$Q$-quantized} if it's both $Q$-quantized by weights and $Q$-quantized by activations.
Note that we do not require the biases to be quantized, but just the linear part of the affine map.
Imposing the quantization constraint to the weights but not to the biases might seem a specious choice, but it is motivated by practical reasons.
First, in most DNNs, the number of inbound synapses (and therefore of weights) of each neuron is usually in the hundreds, implying that the number of weights is two to three orders of magnitudes larger than the number of biases.
Considering that real-valued parameters are usually represented by 32-bits floating-point numbers, quantizing the weights in such a way that each weight can be represented using only one or two bits accounts for almost all the reduction in the storage required by the program's data.
Second, biases are additive terms, whereas weights are multiplicative terms.
In digital hardware, performing multiplications between floating-point numbers is more costly (in terms of energy) than performing multiplications between integers, and performing multiplications between numbers in a given data type is usually more costly than performing additions between numbers in the same data type.
Therefore, quantizing the weights using specific quantization sets allows to replace floating-point multiplications with integer multiplications, or even to replace multiplications with additions if the weights take values in the set $Q = \{ -1, 0, 1 \}$.
It is possible that the quantization set used to quantize the weights is different from the quantization set used to quantize the activations.
In these cases, or in the cases where the set $Q$ could be inferred from the context, we will simply say that the layer map is \textit{quantized}.

We say that a simple feedforward network is a \textbf{quantized neural network} (QNN) if all the layer maps except for the last one are quantized.

\subsection{Layer-wise regularisation}
Many results in the present work involve sequences of regularised functions.
Our use of the term \textit{regularisation} should be intended in functional sense; e.g., as the application of convolution-like operations aimed at smoothing non-differentiabilities or discontinuities.
It should not be confused with the common use of the term in the machine learning community, where it indicates any process which can avoid the so-called \textit{overfitting} problem.

In particular, we focus on layer-wise regularisations: we are interested in approximating a composition \eqref{eq:deep_neural_network} of non-differentiable layer maps \eqref{eq:layer_map} with compositions of differentiable layer maps
\begin{equation*}
    \overline{\Phi} = \overline{\ph}^{L} \circ \dots \circ \overline{\ph}^{1} \,,
\end{equation*}
instead of searching for global regularisations of the network.
Hence, in this work, we will focus on \textbf{compositional convergence} results, by this expression meaning results of the type
\begin{equation*}
    \overline{\Phi}^{k} = \overline{\ph}^{L, k} \circ \dots \circ \overline{\ph}^{1, k} \xrightarrow[k \to +\infty]{} \Phi \,,
\end{equation*}
where the convergence can be in the uniform or pointwise sense.

\subsubsection{Stochastic regularisation of layer maps}
\label{subsub:stochastic}
A first approach to regularisation is taking expectations over families of stochastic functions.
Even though this is substantially equivalent to regularisation by convolution, a reason for adopting such a formal setting is the probabilistic interpretation of the quantitative approximation estimate provided by Theorem~\ref{th:uniform_compositional_approximation}.

For each $\l = 1, \dots, L$, we denote by $(\Xi^{\l}, \Sigma^{\l})$ a measurable space, and by $\mu^{\l}$ a probability measure over it.
Consider layer maps of the form \eqref{eq:layer_map}, where we set $M^{\l} \coloneqq \Xi^{\l}$:
\begin{equation*}
\begin{split}
    \ph_{\xx^{\l}} \,:\,
    X^{\l-1} &\to X^{\l} \\
    \x^{\l-1} &\mapsto \ph_{\xx^{\l}}(\x^{\l-1}) \coloneqq \ph^{\l}(\x^{\l-1}, \xx^{\l}) \,.
\end{split}
\end{equation*}
We call these maps \textbf{stochastic layer maps}.
It is natural to introduce the \textbf{expected layer map}
\begin{equation}\label{eq:expected_layer_map}
\begin{split}
    \overline{\ph}^{\l} \,:\,
    X^{\l-1} &\to X^{\l} \\
    \x^{\l-1} &\mapsto \overline{\ph}^{\l}(\x^{l-1}) \coloneqq \E_{\mu^{\l}}[\ph(\x^{\l-1}, \xx^{\l})] \,.
\end{split}
\end{equation}
In the following, we will sometimes use the notation $\E_{\mu^{\l}}[\ph_{\xx^{\l}}]$ to refer to $\overline{\ph}^{\l}$.

Consistently with \eqref{eq:composition_of_layer_maps} and \eqref{eq:composition_of_regularised_layer_maps}, we define
\begin{align}\label{eq:composition_of_stochastic_layer_maps}
    \overline{\Phi}^{1} &\coloneqq \overline{\ph}^{1} \,, \nonumber \\
    \overline{\Phi}^{\l} &\coloneqq \overline{\ph}^{\l} \circ \overline{\Phi}^{\l-1} \,,\, \l = 2, \dots, L \,.
\end{align}
to denote the compositions of regularised layer maps; we reserve $\overline{\Phi} \coloneqq \overline{\Phi}^{L}$.

\subsubsection{Parametric regularisation of layer maps}
\label{subsub:parametric}
Consider a non-differentiable layer map \eqref{eq:layer_map_decomposition}.
Since $A_{\m}$ is affine, the non-differentiability must be due to the non-linear vector function $\bm{\sigma}$.
Denote by $\sigma$ the non-differentiable activation function on which $\bm{\sigma}$ is built.
Given such a $\sigma$ and a real \textbf{regularisation parameter} $\lambda > 0$, we define the \textbf{regularised activation function} to be any function
\begin{equation}\label{eq:regularised_activation_function}
    \sigma_{\lambda} \,:\, \R \to \R
\end{equation}
such that $\sigma_{\lambda} \in C^{1}(\R)$ and
\begin{equation*}
    \lim_{\lambda \to 0} \sigma_{\lambda}(s) = \sigma(s) \,,\, \forall\, s \in \R
\end{equation*}
(i.e., the regularised activation function converges to the original activation function in pointwise sense).
Consequently, we define the \textbf{regularised layer map} as
\begin{equation}\label{eq:regularised_layer_map}
    \overline{\ph} = \ph_{\lambda, \m} \coloneqq \bm{\sigma}_{\lambda} \circ A_{\m} \,,
\end{equation}
where $\bm{\sigma}_{\lambda}$ is the component-wise application of \eqref{eq:regularised_activation_function}.

Analogously to \eqref{eq:composition_of_layer_maps}, we give the following recursive definition based on \eqref{eq:regularised_layer_map}.
Let $\lambda^{\l} \,,\, \l = 1, \dots, L$, be positive real parameters.
Suppose $\hm^{L} = (\m^{1}, \m^{2}, \dots, \m^{L}) \in \hM^{L}$ is given and set $\hat{\lambda}^{\l} = (\lambda^{1}, \dots, \lambda^{\l})$.
Then we define
\begin{align}
    \Phi_{\hat{\lambda}^{1}, \hm^{1}} &\coloneqq \ph_{\lambda^{1}, \m^{1}} \,, \nonumber \\
    \Phi_{\hat{\lambda}^{\l}, \hm^{\l}} &\coloneqq \ph_{\lambda^{\l}, \m^{\l}} \circ \Phi_{\hat{\lambda}^{\l-1}, \hm^{\l-1}} \,,\, \l = 2, \dots, L \,. \label{eq:composition_of_regularised_layer_maps}
\end{align}

To express the interdependence of different layer maps regularisations, we will define the regularisation parameters $\lambda^{\l} \coloneqq \lambda^{\l}(\lambda) > 0$ as functions of a common real parameter $\lambda > 0$, such that $\lambda^{\l} \to 0$ as $\lambda \to 0$.
In this case, we will define $\Phi_{\lambda, \hm^{\l}} \coloneqq \Phi_{\hat{\lambda}^{\l}, \hm^{\l}}$ and $\Phi_{\lambda} \coloneqq \Phi_{\hat{\lambda}^{L}, \hm^{L}}$.

\bigskip
Some regularisations can be both parametric and stochastic, as shown by the following example.
\begin{xmp}
Let $W = B = \R$, and set $M = \Xi = W \times B$.
Fix $\m = (w, b)$, $\lambda > 0$ and consider the probability measure over $\Xi$ defined as
\begin{equation*}
    \mu \coloneqq \delta_{w} \times \mu_{B} \,,
\end{equation*}
where $\delta_{w}$ is the measure on $W$ concentrated at $w$ and $\mu_{B}$ is the measure associated with a logistic distribution having mean $b + \lambda$ and variance $\lambda^{2}$.
Define the parameter $\xx \coloneqq (w, \beta)$, where $\beta = \beta(\lambda)$ is a random variable distributed according to $\mu_{B}$.
Consider now the stochastic layer map
\begin{equation*}
    \ph_{\xx} \coloneqq \sigma_{\beta} \circ A_{\m} \,,
\end{equation*}
where
\begin{equation*}
\begin{split}
    A_{\m} \,:\,
    \R &\to \R \\
    x &\mapsto s \coloneqq x w + b
\end{split}
\qquad\text{and}\qquad
\begin{split}
    \sigma_{\beta} \,:\,
    \R &\to \R \\
    s &\mapsto H^{+}(s + (\beta(\lambda) - b)) \,,
\end{split}
\end{equation*}
$H^{+}$ being the Heaviside function \eqref{eq:heaviside_zero2one}.
We can define
\begin{equation*}
\begin{split}
    \sigma_{\lambda}(s)
    &\coloneqq \E_{\mu_{B}}[\sigma_{\beta}(s)] \\
    &= \frac{1}{1 + e^{-\frac{s + \lambda}{\lambda^{2}}}} \,.
\end{split}
\end{equation*}
\end{xmp}
\begin{figure}
    \centering
    \includegraphics[width=0.7\textwidth]{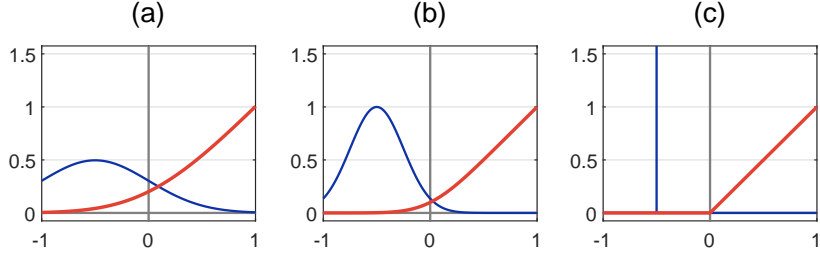}
    \caption{The ReLU function can be smoothed by adding Gaussian noise to its input and applying the expectation operator (a).
        As the variance of the noise diminishes (b), the Gaussian converges (in the weak-$*$ sense) to a Dirac's delta (c), and the regularised functions converge to the ReLU.}\label{fig:relu_smoothing}
\end{figure}

\section{QNNs are as expressive as classical DNNs}
\label{sec:expressivity}
Classical expressivity results for DNNs \cite{Cybenko1989, Hornik1991} assume continuous-valued parameters.
In this section, we prove that this continuity assumption on the parameters is not necessary: QNNs can approximate the same function classes as DNNs that use continuous-valued parameters and activations.

We refer the reader to Section~\ref{sub:qnns} for the notation about QNNs.

\begin{lem}\label{th:expressivity_characteristic_function}
Let $n_{0} > 0$ be an integer, and let $I_{1}, I_{2}, \dots, I_{n_{0}}$ be bounded intervals in $\R$.
Let $P = I_{1} \times I_{2} \times \dots \times I_{n_{0}} \subset \R^{n_{0}}$ be the associated hyperbox.

There exists a QNN with weights in $\{ -1, 0, 1 \}$ and representations in $\{ 0, 1 \}$ that coincides with the characteristic function $\chi_{P}(\x^{0})$.

\begin{proof}
The hyperbox $P$ is the intersection of $n_{0}$-dimensional hyperstripes (i.e., regions delimited by two parallel hyperplanes):
\begin{equation*}
    P = \bigcap_{i=1}^{n_{0}} \{ \x^{0} \,|\, x^{0}_{i} \in I_{i} \} \,,
\end{equation*}
where $x^{0}_{i}$ is the $i$-th component of $\x^{0}$.
In turn, hyperstripes are intersections of half-spaces.
Letting $p_i = \inf I_i$ and $q_i = \sup I_i$, and depending on whether the extremes do or do not belong to the intervals, we have
\begin{align}
    \Big\{ \x^{0} \,|\, x^{0}_{i} \in I_{i} = [p_{i},q_{i}] \Big\} &= \Big\{ \x^{0} \,:\, H^{+}(x^{0}_{i} - p_{i}) = H^{+}(-x^{0}_{i} + q_{i}) = 1)\Big\} \nonumber \\
    \Big\{ \x^{0} \,|\, x^{0}_{i} \in I_{i} = [p_{i},q_{i}) \Big\} &= \Big\{ \x^{0} \,:\, H^{+}(x^{0}_{i} - p_{i}) = H^{-}(-x^{0}_{i} + q_{i}) = 1)\Big\} \nonumber \\
    \Big\{ \x^{0} \,|\, x^{0}_{i} \in I_{i} = (p_{i},q_{i}] \Big\} &= \Big\{ \x^{0} \,:\, H^{-}(x^{0}_{i} - p_{i}) = H^{+}(-x^{0}_{i} + q_{i}) = 1)\Big\} \label{eq:halfspaces_equations} \\
    \Big\{ \x^{0} \,|\, x^{0}_{i} \in I_{i} = (p_{i},q_{i}) \Big\} &= \Big\{ \x^{0} \,:\, H^{-}(x^{0}_{i} - p_{i}) = H^{-}(-x^{0}_{i} + q_{i}) = 1)\Big\} \,, \nonumber
\end{align}
where $H^{+}$ and $H^{-}$ are the Heaviside functions defined in \eqref{eq:heaviside_zero2one} and \eqref{eq:heaviside_zero2zero}.

Let us define the network
\begin{equation*}
    \Phi(\x^{0}) \coloneqq \sigma^{2}\left( \langle \bm{\sigma}^{1}(\x^{0} \W^{1} + \b^{1}), \w^{2} \rangle + b^{2} \right) \,,
\end{equation*}
where $\W^{1} \in \{ -1, 0, 1 \}^{n_{0} \times 2n_{0}} \,,\, \b^{1} \in \R^{1 \times 2n_{0}} \,,\, \w^{2} = \mathbf{1}_{2n_{0}}, b^{2} = -2n_{0}, \sigma^{2} = H^{+}$ and
\begin{equation*}
    \bm{\sigma}^{1}(\x^{0} \W^{1} + \b^{1}) =
    \begin{pmatrix}
        \sigma^{1}_{1}\left( \sum_{i=1}^{n_{0}} x^{0}_{i} w^{1}_{i,1} + b^{1}_{1} \right) \\
        \sigma^{1}_{2}\left( \sum_{i=1}^{n_{0}} x^{0}_{i} w^{1}_{i,2} + b^{1}_{2} \right) \\
        \dots \\
        \sigma^{1}_{2n_{0}}\left( \sum_{i=1}^{n_{0}} x^{0}_{i} w^{1}_{i,2n_{0}} + b^{1}_{2n_{0}} \right)
    \end{pmatrix}^{T}
\end{equation*}
has components defined by
\begin{equation*}
    \sigma^{1}_{i} \coloneqq
    \begin{cases}
        H^{+} \,, &\text{if } p_{i} \in I_{i} \,, \\
        H^{-} \,, &\text{if } p_{i} \notin I_{i} \,,
    \end{cases}
    \quad \text{ and } \quad
    \sigma^{1}_{i+n_{0}} \coloneqq
    \begin{cases}
        H^{+} \,, &\text{if } q_{i} \in I_{i} \,, \\
        H^{-} \,, &\text{if } q_{i} \notin I_{i} \,,
    \end{cases}
\end{equation*}
for $i = 1, 2, \dots, n_{0}$.
If we denote the Kronecker delta by $\delta_{i,j}$, $\W^{1}, \b^{1}$ are defined as
\begin{equation*}
    W^{1}_{i,j} \coloneqq
    \begin{cases}
        \delta_{i,j} \,, &\text{if } 1 \le j \le n_{0} \,, \\
        -\delta_{i,(j-n_{0})} \,, &\text{if } n_{0} + 1 \le j \le 2n_{0} \,,
    \end{cases}
    \quad \text{ and } \quad
    b^{1}_{j} \coloneqq
    \begin{cases}
        -p_{j}, &\text{if } 1 \le j \le n_{0} \,, \\
        q_{j-n_{0}}, &\text{if } n_{0}+1 \le j \le 2n_{0} \,,
    \end{cases}
\end{equation*}
for $i = 1, 2, \dots, n_{0}$ and $j = 1, 2, \dots, 2n_{0}$.

We now observe that $\Phi(\x^{0}) = 1$ if and only if the argument of the outer activation $\sigma^{2}$ is non-negative, i.e., if
\begin{equation*}
    \langle \bm{\sigma}^{1}(\x^{0} \W^{1} + \b^{1}), \w^{2} \rangle \ge -b^{2} \,.
\end{equation*}
Since we defined $b^{2} = -2n_{0}$, this can be satisfied if and only if equality holds:
\begin{equation*}
    \bm{\sigma}^{1}(\x^{0} \W^{1} + \b^{1}) = \w^{2} \,.
\end{equation*}
The latter vector equation corresponds to satisfying all the systems
\begin{equation*}
    \begin{cases}
        \sigma^{1}_{j} \left( \sum_{i=1}^{n_{0}} x^{0}_{i} w^{1}_{i,j} + b^{1}_{j} \right) = 1 \,, \\
        \sigma^{1}_{j+n_{0}} \left( \sum_{i=1}^{n_{0}} x^{0}_{i} w^{1}_{i,(j+n_{0})} + b^{1}_{j+n_{0}} \right) = 1 \,,
    \end{cases}
    \, j = 1, \dots, n_{0} \,,
\end{equation*}
which are equivalent, by the definitions of $W^{1}_{i,j}, b^{1}_{j}, \sigma^{1}_{j}$, to the systems of equations involving $H^+,H^-$, which appear in \eqref{eq:halfspaces_equations}.
This amounts to requiring that $\x^{0} \in \{ \x^{0} \,|\, x^{0}_{i} \in I_{i} \} \,,\, \,\forall\, i = 1, \dots, n_{0}$, and thus that $\x^{0} \in P$.
Hence $\Phi(\x^{0}) = \chi_{P}(\x^{0})$.
\end{proof}

\begin{figure}
    \centering
    \includegraphics[width=0.4\textwidth]{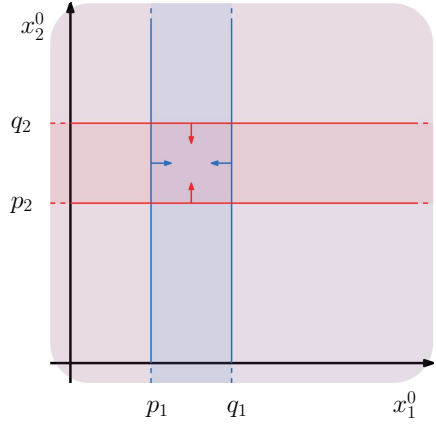}
    \caption{When $n_{0} = 2$ (i.e., when we are in the real plane), the hyperbox $P = [p_{1}, q_{1}] \times [p_{2}, q_{2}]$ is the intersection of four half-spaces.
        Algebraically, it can be expressed as the set $\{ \x^{0} \in \R^{2} \,|\, H^{+}(x^{0}_{1} - p_{1}) + H^{+}(-x^{0}_{1} + q_{1}) + H^{+}(x^{0}_{2} - p_{2}) + H^{+}(-x^{0}_{2} + q_{2}) \geq 4 \}$.}\label{fig:expressivity_characteristic_function}
\end{figure}
\end{lem}

\begin{thm}[Uniform approximation by QNNs]\label{th:expressivity}
Let $X^{0} \coloneqq [0, S]^{n_{0}} \subset \R^{n_{0}}$.
Let $\Lambda > 0$ be a positive real number, and denote by $\Lip_{\Lambda}(X^{0})$ the class of bounded functions $f \,:\, X^{0} \to \R$ with Lipschitz norm bounded by $\Lambda$.

Then, for every $f \in \Lip_{\Lambda}(X^{0})$ and $\eps > 0$, there exists a quantized neural network
\begin{equation*}
    \Phi \coloneqq \ph_{\m^{3}} \circ \ph_{\m^{2}} \circ \ph_{\m^{1}} \,:\, X^{0} \to \R \,,
\end{equation*}
such that
\begin{equation*}
    \| \Phi - f \|_{L^{\infty}(X^{0})} \coloneqq \sup_{\x \in X^{0}} \big| \Phi(\x) - f(\x) \big| < \eps \,.
\end{equation*}
In particular, the layer maps $\ph_{\m^{1}}$ and $\ph_{\m^{2}}$ have quantized weights $\W^{1} \in \{ -1, 0, 1 \}^{n_{0} \times n_{1}}, \W^{2} \in \{ -1, 0, 1 \}^{n_{1} \times n_{2}}$ (where $n_{1} = n_{1}(\eps)$ and $n_{2} = n_{2}(\eps)$) and $\{ 0, 1 \}$-quantized activation functions, while the layer $\ph_{\m^{3}}$ is a linear map which uses continuous-valued parameters.
Moreover, the number of neurons required by $\Phi$ to reach the given approximation degree $\eps$ is bounded by
\begin{equation}\label{eq:model_size_bound}
    (2n_{0} + 2) \left\lceil \frac{\Lambda \sqrt{n_{0}} S}{\eps} \right\rceil ^{n_{0}} \,.
\end{equation}

\begin{proof}
The proof is composed of two parts:
\begin{enumerate}
    \item\label{th:expressivity_step1} first, we explicitly construct a QNN that can exactly represent a function $f$ which is constant on hyperboxes;
    \item\label{th:expressivity_step2} then, we show that for any function $f \in \Lip_{\Lambda}(X_{0})$, there exists a simple function defined on hyperboxes (i.e., a function which is constant on hyperboxes) which approximates $f$ arbitrarily well.
\end{enumerate}

\textit{Step \ref{th:expressivity_step1}}.
Let $N$ be a positive integer.
Let $\{ P_{1}, P_{2}, \dots, P_{N} \}$ be a family of hyperboxes such that $X^{0} = \bigcup_{s=1}^{N} P_{s}$ and $P_{s_{1}} \cap P_{s_{2}} = \emptyset \,,\, \,\forall\, s_{1}, s_{2} \in \{ 1, 2, \dots, N \} \,,\, s_{1} \neq s_{2}$; i.e., $\{P_{s}\}_{s = 1, \dots N}$ is a partition of $X^{0}$.
We define the simple function
\begin{equation}\label{eq:expressivity_simple_function}
\begin{split}
    \hat{f} \,:\,
    \R^{n_{0}} &\to \R \\
    \x^{0} &\mapsto \sum_{s=1}^{N} f_{s} \chi_{P_{s}}(\x^{0}) \,,
\end{split}
\end{equation}
where $f_{s} \in \R, s = 1, 2, \dots, N$, which is constant on hyperboxes.

Set now $n_{1} \coloneqq 2n_{0} N$ and define the block matrix
\begin{equation*}
    \W^{1} = (\W^{1,1}, \W^{1,2}, \dots, \W^{1,N})
\end{equation*}
as the concatenation of matrices $\W^{1,s} \in \{ -1, 0, 1 \}^{n_{0} \times 2n_{0}} \,,\, s = 1, 2, \dots, N$, and the block row vector
\begin{equation*}
    \b = (\b^{1,1}, \b^{1,2}, \dots, \b^{1,N})
\end{equation*}
as the concatenation of row vectors $\b^{1,s} \in \R^{1 \times 2n_{0}} \,,\, s = 1, \dots, N$.
We define the first layer as
\begin{equation*}
\begin{split}
    \ph_{\m^{1}} \,:\,
    \R^{n_{0}} &\to \{ 0, 1 \}^{n^{1}} \\
    \x^{0} &\mapsto \left( \ph_{(\W^{1,1}, \b^{1,1})}(\x^{0}), \ph_{(\W^{1,2}, \b^{1,2})}(\x^{0}), \dots, \ph_{(\W^{1,N}, \b^{1,N})}(\x^{0}) \right) \,,
\end{split}
\end{equation*}
where $\ph_{(\W^{1,s}, \b^{1,s})} \,:\, \R^{n_{0}} \to \{ 0, 1 \}^{2n_{0}}$ is the one-layer ternary network measuring the membership of $\x^{0}$ to the half-spaces enclosing the hyperbox $P_{s}$ (i.e., each $\ph_{\W^{1,s}, \b^{1,s}}$ is an instance of the first layer map of the network described in Lemma~\ref{th:expressivity_characteristic_function}).
Each instance is applied to a different $P_{s}$, in parallel with the others.

Set now $n_{2} \coloneqq N$, and define the block matrix
\begin{equation*}
    \W^{2} = (\w^{2,1}, \w^{2,2}, \dots, \w^{2,N})
\end{equation*}
where $\w^{2, s} \in \{ 0, 1 \}^{n_{1} \times 1}$ are the column vectors
\begin{equation*}
    w^{2,s}_{i} \coloneqq
    \begin{cases}
        1 \,, &\text{if } 2n_{0}(s-1) + 1 \leq i \leq 2n_{0}s \,, \\
        0 \,, &\text{otherwise} \,,
    \end{cases}
    \, s = 1, 2, \dots, N \,
\end{equation*}
and $\b^{2} \in \R^{1 \times N}$ is a row vector with all equal components:
\begin{equation*}
    b^{2}_{s} = -2n_{0} \,,\, s = 1, 2, \dots, N \,.
\end{equation*}
We define the second layer as
\begin{equation*}
\begin{split}
    \ph_{\m^{2}} \,:\,
    \{ 0, 1 \}^{n_{1}} &\to \{ 0, 1 \}^{n_{2}} \\
    \x^{1} &\mapsto \bm{\sigma}(\x^{1} \W^{2} + b^{2}) \,,
\end{split}
\end{equation*}
where $\bm{\sigma}$ is the component-wise application of \eqref{eq:heaviside_zero2one}.
The map $\ph_{\m^{2}} \circ \ph_{\m^{1}}$ thus measures (in parallel) the membership of a point $\x^{0}$ to all the hyperboxes $P_{s}$.
Since $\{ P_{s} \}_{s = 1, 2, \dots N}$ is a partition of $X^{0}$, just one neuron of the second layer can be active at a time when processing a given input $\x^{0} \in X^{0}$ (i.e., only one neuron can have value $1$).

Finally, we define the column vector $\w^{3} \in \R^{N}$ with components
\begin{equation*}
    w^{3}_{s} \coloneqq f_{s} \,,\, s = 1, 2, \dots, N \,,
\end{equation*}
where $f_{s}$ is the value of $\hat{f}$ on $P_{s}$.
We define the linear map
\begin{equation*}
\begin{split}
    \ph_{\m^{3}} \,:\,
    \{ 0, 1 \}^{N} &\to \R \\
    \x^{2} &\mapsto \langle \x^{2}, \w^{3} \rangle \,.
\end{split}
\end{equation*}

The map
\begin{equation*}
    \Phi \coloneqq \ph_{\m^{3}} \circ \ph_{\m^{2}} \circ \ph_{\m^{1}}
\end{equation*}
is the desired QNN representing \eqref{eq:expressivity_simple_function} exactly.

\medskip

\textit{Step \ref{th:expressivity_step2}}.
Let $f\in \Lip_{\Lambda}(X^{0})$ and $\eps > 0$ be fixed.

Let $n$ be an integer that satisfies
\begin{equation*}
    n \ge \frac{\Lambda \sqrt{n_{0}} S}{\eps} \,,
\end{equation*}
and set $N \coloneqq n^{n_{0}}$.
Consider the family of hypercubes $P_{s}$ with side length $\delta = S/n$, forming a partition $\{ P_{s} \}_{s = 1, 2, \dots, N}$ of $X^{0}$.
For each $s = 1, 2, \dots, N$ we can identify the hypercube $P_{s}$ by the index tuple $(i^{s_{0}}, \dots, i^{s_{n_{0}-1}})$ whose $n_{0}$ components are the unique integers $i^{s_{k}}\in \{ 0, 1, \dots, n_{0}-1 \}$ such that
\begin{equation*}
    s-1 = i^{s_{0}} + i^{s_{1}} n + i^{s_{2}} n^{2} + \dots + i^{s_{n_{0}-1}} n^{n_{0}-1} \,.
\end{equation*}
Then, the hypercube $P_{s}$ is given by
\begin{equation*}
  P_{s} = I_{s_{0}} \times I_{s_{1}} \times \dots \times I_{s_{n_{0}-1}}
\end{equation*}
where
\begin{equation*}
    I_{s_{k}} =
    \begin{cases}
        [i^{s_{k}}\delta, (i^{s_{k}}+1)\delta) \,, &\text{if } 0 \leq s_{k} < n_{0} - 1 \,, \\
        [i^{s_{k}}\delta, (i^{s_{k}}+1)\delta] \,, &\text{if } s_{k} = n_{0} - 1 \,.
    \end{cases}
\end{equation*}

Now, define $\hat{f}$ as in \eqref{eq:expressivity_simple_function} by setting $f_{s}$ to be the integral average of $f$ on $P_{s}$ for each $s = 1, 2, \dots, N$.
By the previous step, we know that this function can be represented exactly by a QNN.
Hence, we are left with showing that
\begin{equation*}
    \big| \hat{f}(\x^{0}) - f(\x^{0}) \big| \le \eps \,,\, \,\forall \, \x^{0} \in X^{0} \,.
\end{equation*}
For every given $\x^{0} \in X^{0}$, there exists $s \in \{ 1, 2, \dots, N \}$ such that $\x^{0} \in P_{s}$ (since $\{ P_{s} \}_{s = 1, 2, \dots, N}$ is a partition of $X^{0}$).
By definition of integral average and by the continuity of $f$, there exists $\x^{0}_{s} \in P_{s}$ such that $\hat{f}(\x^{0}) \coloneqq f_{s} = f(\x^{0}_{s})$.
Hence, by the Lipschitz property of $f$, we obtain
\begin{equation*}
    \big| \hat{f}(\x^{0}) - f(\x^{0}) \big|
    = \big| f(\x^{0}_{s}) - f(\x^{0}) \big| 
    \leq \lambda \big| \x^{0}_{s} - \x^{0} \big| 
    \leq \Lambda \diam(P_{s}) 
    = \Lambda \frac{\sqrt{n_{0}}S}{n}
    < \eps \,.
\end{equation*}
Since this estimate holds for any $\x^{0} \in X^{0}$, the proof is complete.
\end{proof}

\begin{figure}
    \centering
    \includegraphics[width=\textwidth]{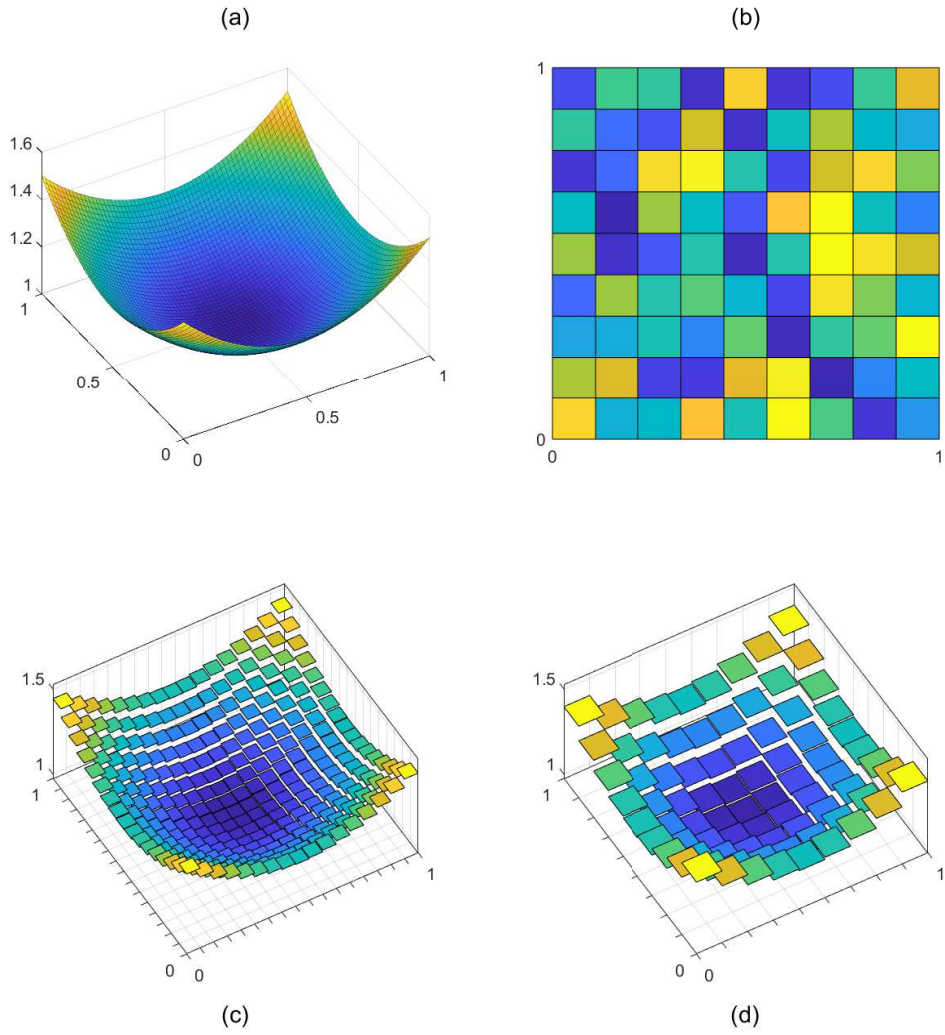}
    \caption{The function $f(x, y) = (x - 1/2)^{2} + (y - 1/2)^{2} +1$ defined over the domain $D = [0, 1]^{2}$  (a) is approximated at precision $\eps = 1/8$ (c) and at precision $\eps = 1/4$ (d) by different QNNs.
        Sub-figure (b) shows the partition of the domain sufficient to obtain the approximation shown in (d).}\label{fig:expressivity}
\end{figure}
\end{thm}

\begin{rmk}
Since $\bigcup_{\Lambda>0} \Lip_{\Lambda}(X^{0})$ is dense in $C^{0}(X^{0})$, we infer from Theorem~\ref{th:expressivity} that QNNs are essentially as expressive as classical networks.
The theorem implies that the accuracy gaps observed experimentally between QNNs and DNNs using continuous-valued parameters are not intrinsic to the class of quantized networks.
Notice moreover that, in order to approximate generic real-valued functions, the last layer must be parametrised by continuous variables.
\end{rmk}

\begin{rmk}
In most applications, the relationship described in \eqref{eq:model_size_bound} between the dimension of the input space $n_{0}$, the degree of approximation $\eps$, and the Lipschitz constant $\Lambda$ gives an impractically loose bound on the model size.
But it is worth considering that this bound could be made tighter, e.g., by applying some feature extraction process aimed at reducing $n_{0}$, such as principal component analysis (PCA), kernel PCA or manifold learning techniques.
Interestingly, existing experimental research in QNNs has often focused on networks whose first layers used full-precision parameters \cite{Rastegari2016, Zhou2016, Zhu2017}.
An intuitive explanation of these choices is that full-precision initial layers might be able to learn features that are more suitable for the successive processing performed by QNNs.
\end{rmk}

\section{Solving the training problem for non-differentiable DNNs}
\label{sec:non-differentiability}
Despite the practical effectiveness of the ReLU activation function, its derivative is not defined at the origin, which is a formal inconsistency for training algorithms based on gradient-descent.
In this section, we show how a network using Lipschitz activation functions can be interpreted as the limit of a sequence of more regular (possibly differentiable) networks.
Consequently, we can see the training problem for a ReLU network as the limit of a sequence of well-defined training problems which can be correctly solved by gradient descent.

The following theorem provides an explicit bound on the $L^{\infty}$ distance between a continuous feedforward network \eqref{eq:deep_neural_network} and a corresponding composition of stochastically regularised layer maps \eqref{eq:expected_layer_map}.
We refer the reader to Section~\ref{subsub:stochastic} for the notation.

\begin{thm}[Uniform compositional approximation]\label{th:uniform_compositional_approximation}
Let $L$ be a positive integer and let $X^{\l-1}$ and $\Xi{^\l}$ be compact subsets of some given Euclidean spaces, for $\l = 1, \dots, L$.
Assume that each map $\ph_{\xx^{\l}}(\x^{\l-1}) = \ph^{\l}(\x^{\l-1}, \xx^{\l})$ is continuous in both variables $\x^{\l-1}$ and $\xx^{\l}$, where $\xx^{\l}$ is a random variable distributed according to a measure $\mu^{\l}$, and denote by $\eta^{\l}$ its modulus of continuity, that is, $\eta^{\l} \,:\, [0, +\infty) \to [0, +\infty)$ is continuous, strictly increasing, and satisfies 
\begin{equation*}
    \eta^{\l}(0) = 0\,,\qquad \|\ph^{\l}(\x, \xx) - \ph^{\l}(\x', \xx')\| \le \eta^{\l}(\|\x - \x'\| + \|\xx - \xx'\|)\quad\forall\, (\x, \xx),\ (\x', \xx')\in X^{\l-1} \times \Xi^{\l}\,.
\end{equation*}
For $\l=1,\dots,L$ we define $\theta^\l$ by induction as
\begin{align*}
    \theta^{1} &\coloneqq \int_{\Xi^{1}} \eta^{1}(\| \xx^{1} - \m^{1} \|) d\mu^{\l}(\xx^{1}) \,, \\
    \theta^{\l} &\coloneqq \int_{\Xi^{\l}} \eta^{\l}(\| \xx^{\l} - \m^{\l} \|) d\mu^{\l}(\xx^{\l}) + \eta^{\l}(\theta^{\l-1}) \,,\, \l = 2, \dots, L \,.
\end{align*}

Then we have
\begin{equation}\label{eq:uniform_compositional_approximation_theta_estimate}
    \| \overline{\Phi}^{\l} - \Phi_{\hm^{\l}} \|_{L^{\infty}(X^{0})} \leq \theta^{\l} \,,\, \,\forall\, \l = 1, \dots, L \,.
\end{equation}
In particular, when the layer maps $\ph^\l(\x, \xx)$ are Lipschitz, we have
\begin{equation}\label{eq:uniform_compositional_approximation_lipschitz_estimate}
    \| \overline{\Phi}^{\l} - \Phi_{\hm^{\l}} \|_{L^{\infty}(X^{0})} \le \sum_{i=1}^{\l} \left(\prod_{j=i}^{\l} C_{j}\right) \, {\text{Var}}_{1}(\xx^{i}) \,,
\end{equation}
where ${\text{Var}}_{1}(\xx^{i})$ denotes the $L^{1}$ variance of $\xx^{i}$, and $C_{j}$ is the Lipschitz constant of $\ph^{j}$.

\begin{proof}
We prove the theorem by induction.
Fix $\x^{0} \in X^{0}$.

The base step is immediate:
\begin{equation*}
\begin{split}
    \| \overline{\Phi}^{1}(\x^{0}) - \Phi_{\hm^{1}}(\x^{0}) \|
    &= \Big\| \int_{\Xi^{1}} \left( \ph^{1}(\x^{0}, \xx^{1}) - \ph^{1}(\x^{0}, \m^{1}) \right) d\mu^{1}(\xx^{1}) \Big\| \\
    &\leq \int_{\Xi^{1}} \left\| \ph^{1}(\x^{0}, \xx^{1}) - \ph^{1}(\x^{0}, \m^{1}) \right\| d\mu^{1}(\xx^{1}) \\
    &\leq \int_{\Xi^{1}} \eta^{\l}(\| \xx^{1} - \m^{1} \|)\, d\mu^{1}(\xx^{1}) \\
    &= \theta^{1} \,.
\end{split}
\end{equation*}

We now assume that
\begin{equation*}
    \| \overline{\Phi}^{\l}(\x^{0}) - \Phi_{\hm^{\l}}(\x^{0}) \| \leq \theta^{\l}
\end{equation*}
holds; then, our goal is to show that
\begin{equation}\label{eq:uniform_compositional_approximation_inductive_thesis}
    \| \overline{\Phi}^{\l+1}(\x^{0}) - \Phi_{\hm^{\l+1}}(\x^{0}) \| \leq \theta^{\l+1} \,.
\end{equation}
First, by the triangular inequality we estimate the first term as follows:
\begin{equation*}
\begin{split}
    \| \overline{\Phi}^{\l+1}(\x^{0}) - \Phi_{\hm^{\l+1}}(\x^{0}) \|
    &= \| \overline{\ph}^{\l+1}(\overline{\Phi}^{\l}(\x^{0})) - \ph_{\m^{\l+1}}(\Phi_{\hm^{\l}}(\x^{0})) \| \\
    &= \Big\| \int_{\Xi^{\l+1}} \left( \ph^{\l+1}(\overline{\Phi}^{\l}(\x^{0}), \xx^{\l+1}) - \ph^{\l+1}(\Phi_{\hm^{\l}}(\x^{0}), \m^{\l+1}) \right) d\mu^{\l+1}(\xx^{\l+1}) \Big\| \\
    &\leq \int_{\Xi^{\l+1}} \left\| \ph^{\l+1}(\overline{\Phi}^{\l}(\x^{0}), \xx^{\l+1}) - \ph^{\l+1}(\Phi_{\hm^{\l}}(\x^{0}), \m^{\l+1}) \right\| d\mu^{\l+1}(\xx^{\l+1}) \\
    &\leq \int_{\Xi^{\l+1}} \left\| \ph^{\l+1}(\overline{\Phi}^{\l}(\x^{0}), \xx^{\l+1}) - \ph^{\l+1}(\overline{\Phi}^{\l}(\x^{0}), \m^{\l+1}) \right\| d\mu^{\l+1}(\xx^{\l+1}) + \\
    &\quad\quad + \int_{\Xi^{\l+1}} \left\| \ph^{\l+1}(\overline{\Phi}^{\l}(\x^{0}), \m^{\l+1}) - \ph^{\l+1}(\Phi_{\hm^{\l}}(\x^{0}), \m^{\l+1}) \right\| d\mu^{\l+1}(\xx^{\l+1}) \\
    &=: A + B \,.
\end{split}
\end{equation*}
The term $A$ satisfies
\begin{equation*}
    \int_{\Xi^{\l+1}} \left\| \ph^{\l+1}(\overline{\Phi}^{\l}(\x^{0}), \xx^{\l+1}) - \ph^{\l+1}(\overline{\Phi}^{\l}(\x^{0}), \m^{\l+1}) \right\| d\mu^{\l+1}(\xx^{\l+1}) \leq \int_{\Xi^{\l+1}} \eta^{\l+1}(0 + \| \xx^{\l+1} - \m^{\l+1} \|) d\mu^{\l+1}(\xx^{\l+1}) \,,
\end{equation*}
whereas the term $B$ satisfies
\begin{equation*}
\begin{split}
    &\int_{\Xi^{\l+1}} \left\| \ph^{\l+1}(\overline{\Phi}^{\l}(\x^{0}), \m^{\l+1}) - \ph^{\l+1}(\Phi_{\hm^{\l}}(\x^{0}), \m^{\l+1}) \right\| d\mu^{\l+1}(\xx^{\l+1}) \\
    &\quad\quad\quad\quad \leq \int_{\Xi^{\l+1}} \eta^{\l+1}\Big(\| \overline{\Phi}^{\l}(\x^{0}) - \Phi_{\hm^{\l}}(\x^{0}) \|\Big)\, d\mu^{\l+1}(\xx^{\l+1}) \\
    &\quad\quad\quad\quad = \eta^{\l+1}(\theta^{\l}) \,,
\end{split}
\end{equation*}
which implies \eqref{eq:uniform_compositional_approximation_inductive_thesis} for any given $\x^{0}$.

Being the estimate independent of $\x^{0}$, it holds uniformly on $X^{0}$.
Finally, \eqref{eq:uniform_compositional_approximation_lipschitz_estimate} is an easy consequence of \eqref{eq:uniform_compositional_approximation_theta_estimate} and of the estimate $\eta^{j}(t) \le C_{j} t$ for $t > 0$.
\end{proof}
\end{thm}

The following corollary provides a criterion to iteratively approximate a Lipschitz network \eqref{eq:deep_neural_network}.

\begin{cor}
Let $X^{\l-1}$ and $\Xi{^\l}$ be compact subsets of given Euclidean spaces.
Assume that, for all $\l = 1, \dots, L$, the map $\ph_{\xx^{\l}}(\x^{\l-1}) = \ph^{\l}(\x^{\l-1}, \xx^{\l})$ is continuous in both variables $\x^{\l-1}$ and $\xx^{\l}$, and denote by $\Phi$ their composition, as in \eqref{eq:deep_neural_network}.
For each $\l = 1, \dots, L$, we fix $\m^{\l} \in \Xi^{\l}$ and let $\{{\mu}^{\l}_{k}\}_{k \in \N}$ be a sequence of probability measures on $\Xi^{\l}$ converging in the weak-$*$ sense to the Dirac's delta $\delta_{\m^{\l}}$ as $k \to \infty$.
Then, similarly to \eqref{eq:composition_of_stochastic_layer_maps}, we set
\begin{align*}
    \overline{\Phi}^{1, k} &\coloneqq \overline{\ph}^{1, k} \,, \nonumber \\
    \overline{\Phi}^{\l, k} &\coloneqq \overline{\ph}^{\l, k} \circ \overline{\Phi}^{\l-1, k} \,,\, \l = 2, \dots, L \,,
\end{align*}
where $\overline{\ph}^{\l, k} \coloneqq \E_{\mu^{\l}_{k}}[\ph_{\xx^{\l}}]$.

We have $\ds\lim_{k \to \infty} \overline{\Phi}^{L, k}(\x) = \Phi(\x), \,\forall\, \x \in X^{0}$.

\begin{proof}
Using the terminology of Theorem \ref{th:uniform_compositional_approximation}, we set
\begin{align*}
    \theta^{1, k} &\coloneqq \int_{\Xi^{1}} \eta^{1}(\| \xx^{1} - \m^{1} \|) d\mu^{\l}_{k}(\xx^{1}) \,, \\
    \theta^{\l, k} &\coloneqq \int_{\Xi^{\l}} \eta^{\l}(\| \xx^{\l} - \m^{\l} \|) d\mu^{\l}_{k}(\xx^{\l}) + \eta^{\l}(\theta^{\l-1}_{k}) \,,\qquad \l = 2, \dots, L \,.
\end{align*}
In this way, the proof is reduced to the simple observation which leverages the weak-$*$ convergence of the measures:
\begin{equation*}
    \int_{\Xi^{\l}} \eta^{\l}(\| \xx^{\l} - \m^{\l} \|)\, d\mu^{\l}_{k}(\xx^{\l}) \xrightarrow[k \to \infty]{}  \int_{\Xi^{\l}} \eta^{\l}(\| \xx^{\l} - \m^{\l} \|)\, d\delta_{\m^\l}(\xx^{\l}) = 0\,.
\end{equation*}
\end{proof}
\end{cor}

\begin{rmk}\label{rmk:uniform_compositional_approximation_annealing}
Expression \eqref{eq:uniform_compositional_approximation_lipschitz_estimate} can be used as a valuable tool to derive appropriate \textit{annealing criteria} for choosing the noise variance of each stochastic layer map.
For example, consider the case of transfer learning \cite{Bengio2012}.
In this context, one might be interested not only in controlling the approximation of the global map (i.e., $\| \overline{\Phi}^{L} - \Phi \|_{L^{\infty}(X^{0})}$), but also in guaranteeing that all the intermediate compositions approximate the corresponding representation maps within a given degree of approximation $\eps$ (i.e., $\| \overline{\Phi}^{\l} - \Phi_{\hm^{\l}} \|_{L^{\infty}(X^{0})} < \eps$ for all $\l = 1, \dots, L-1$).
If we assume that $C_{j} \geq 1$ for each $j = 1, \dots, L$ (i.e., the layer maps are non-contractive), to achieve such a condition it is sufficient to impose that each term of the sum appearing on the right-hand side of \eqref{eq:uniform_compositional_approximation_lipschitz_estimate} equally contributes to the whole estimate:
\begin{equation}\label{eq:uniform_compositional_approximation_annealing_noise_bounds}
    {\text{Var}}_{1}(\xx^{i}) \le \left( \ds\prod_{j = i}^{L} C_{j}^{-1} \right) \left( \frac{\eps}{L} \right) \,,\qquad \text{for every } i = 1, \dots, L \,.
\end{equation}
To see why, observe that for any $\hat{\l} = 1, \dots, L$ the following holds:
\begin{equation*}
\begin{split}
    \sum_{i = 1}^{\hat{\l}} \left( \prod_{j = i}^{\hat{\l}} C_{j} \right) {\text{Var}}_{1}(\xx^{i})
    &\leq \sum_{i = 1}^{\hat{\l}} \left( \prod_{j = i}^{\hat{\l}} C_{j} \right) \left( \frac{\eps}{\left( \prod_{j = i}^{L} C_{j} \right) L} \right) \\
    &\leq \sum_{i = 1}^{\hat{\l}} \left( \prod_{j = i}^{\hat{\l}} C_{j} \right) \left( \frac{\eps}{\left( \prod_{j = i}^{\hat{\l}} C_{j} \right) L} \right) \\
    &= \sum_{i = 1}^{\hat{\l}} \left( \frac{\eps}{L} \right) \leq \eps \,.
\end{split}
\end{equation*}
\end{rmk}

\begin{rmk}
If we consider layer maps as in \eqref{eq:layer_map_decomposition} using the ReLU activation function, their Lipschitz constants satisfy $C_{\l} = \| \W^{\l} \|_{2}$ for $\l = 1, \dots, L$, where $\| \W^{\l} \|_{2}$ is the spectral norm of the linear operator associated to the weight matrix $\W^{\l}$.
Since the weights are updated during training, the bounds provided by \eqref{eq:uniform_compositional_approximation_annealing_noise_bounds} can change over time.
Efficient ways to estimate $\| \W^{\l} \|_{2}$ are available in the existing literature \cite{Scaman2018}. One could therefore devise training algorithms that include dynamic updates of the noise-based regularisation.
\end{rmk}

\section{Approximating discontinuous neural networks}
\label{sec:discontinuity}

\subsection{Regularising discontinuous activation functions}
In Section~\ref{sec:non-differentiability} we justified the application of gradient descent algorithms in presence of continuous but non-differentiable activation functions.
It would be very interesting to obtain similar results in the discontinuous case.
Of course, strong estimates like \eqref{eq:uniform_compositional_approximation_theta_estimate} are unlikely to hold, simply because the uniform approximation of a discontinuous function by sequences of continuous functions is not possible.
Nevertheless, some pointwise compositional convergence can still be granted if the discontinuous activation functions are suitably approximated.
To avoid technical complications, we explain this idea in the specific setting of Heaviside activation functions.

Consider a network \eqref{eq:deep_neural_network} parametrised by $\hm^{L} = (\m^{1}, \m^{2}, \dots, \m^{L}) \in \hM^{L}$ and using the Heaviside function $H^{+}$ defined in \eqref{eq:heaviside_zero2one} as its activation function (i.e., $\Phi$ is $\{ 0, 1 \}$-quantized by activations).
According to the notation defined in Section~\ref{subsub:parametric}, let $\lambda > 0$ be a real parameter, and let $\lambda^{\l} \coloneqq \lambda^{\l}(\lambda) > 0 \,,\, \l = 1, \dots, L$ be regularisation parameters that satisfy
\begin{align}
    &\lambda^{\l} \xrightarrow[\lambda \to 0]{} 0 \,; \label{eq:regularisation_parameter_H1} \\
    &\sigma_{\lambda^{\l}}(s) \xrightarrow[\lambda^{\l} \to 0]{} H^+(s) \,,\, \,\forall\, s \in \R \,; \label{eq:regularisation_parameter_H2} \\
    &\text{$\sigma_{\lambda^{\l}}$ is strictly increasing} \,; \label{eq:regularisation_parameter_H3} \\
    &0 \leq \sigma_{\lambda^{\l}}(s) \leq 1 \,,\, \,\forall\, s \in \R \,. \label{eq:regularisation_parameter_H4}
\end{align}

We recall that a \textit{convergence rate} is a continuous, increasing function $r \,:\, R_{0}^{+} \to R_{0}^{+}$ such that $r(\lambda) \to 0$ as $\lambda \to 0$.
\begin{dfn}[Rate convergence]\label{dfn:rate_convergence}
Let $\lambda > 0$ be a real parameter, and define regularisation parameters $\lambda^{\l} \coloneqq \lambda^{\l}(\lambda) > 0 \,, \,\l = 1, \dots, L$ that satisfy condition \eqref{eq:regularisation_parameter_H1}.
Let $\{ \sigma_{\lambda^{\l}} \}_{\l = 1}^{L}$ be a corresponding sequence of regularised Heaviside functions that satisfy conditions \eqref{eq:regularisation_parameter_H2}, \eqref{eq:regularisation_parameter_H3} and \eqref{eq:regularisation_parameter_H4}.
If there exist convergence rates $r^{1}(\lambda), \dots, r^{L}(\lambda)$ such that for every $\eps > 0$ we have
\begin{eqnarray}
\label{th:pointwise_convergence_hp_i} \sigma_{\lambda^{\l}}^{-1}(\eps r^{\l}(\lambda)) \xrightarrow[\lambda \to 0]{} 0\,,& \\
\label{th:pointwise_convergence_hp_ii} \sigma_{\lambda^{\l}}^{-1}(1 - \eps r^{\l}(\lambda)) \xrightarrow[\lambda \to 0]{} 0 \,,& \\
\label{th:pointwise_convergence_hp_iii} \frac{1 - \sigma_{\lambda^{\l}}(0)}{r^{\l}(\lambda)} \xrightarrow[\lambda \to 0]{} 0\,,&
\end{eqnarray}    
for $\l = 1, \dots, L$, and
\begin{equation}\label{th:pointwise_convergence_hp_iv}
    \frac{r^{\l-1}(\lambda)}{\sigma_{\lambda^{\l}}^{-1}(1 - \eps r^{\l}(\lambda))} \xrightarrow[\lambda \to 0]{} 0\,.
\end{equation}
for $\l = 2, \dots, L$, then we say that the sequence $\{ \sigma_{\lambda^{\l}} \}_{\l = 1}^{L}$ is \textbf{rate convergent}.
\end{dfn}

\begin{rmk}
The geometric idea behind properties \eqref{th:pointwise_convergence_hp_i}, \eqref{th:pointwise_convergence_hp_ii} and \eqref{th:pointwise_convergence_hp_iii} is that the graph of $\sigma_{\lambda^{\l}}$ must converge faster to the horizontal parts than to the vertical part of the graph of $H^{+}$ (an example function satisfying these conditions is depicted in Figure~\ref{fig:pointwise_convergence}b).
Instead, \eqref{th:pointwise_convergence_hp_iv} is a \textit{transmission property} ensuring that the convergence of the $(\l-1)$-th function happens in such a way as to guarantee the convergence of the $\l$-th function.
\end{rmk}

\begin{thm}[Pointwise compositional convergence]\label{th:pointwise_convergence}
Consider a network \eqref{eq:deep_neural_network} parametrised by $\hm^{L} = (\m^{1}, \m^{2}, \dots, \m^{L}) \in \hM^{L}$ and using the Heaviside function $H^{+}$ defined in \eqref{eq:heaviside_zero2one} as its activation function.
Let $\lambda^{\l} = \lambda^{\l}(\lambda) > 0 \,,\, \l = 1, \dots, L$ be regularisation parameters that depend on a common parameter $\lambda > 0$.
For $\l = 1, \dots, L$ let $\sigma_{\lambda^{\l}}$ be the regularised Heaviside activation function of the $\l$-th layer.
Assume that the collection $\{ \sigma_{\lambda^{\l}} \}_{\l = 1}^{L}$ is rate-convergent.

Then, for each fixed $\x^{0} \in X^{0}$ we have
\begin{equation}\label{eq:pointwise_convergence_thesis}
    \frac{\| \Phi_{\lambda, \hm^{\l}}(\x^{0}) - \Phi_{\hm^{\l}}(\x^{0}) \|}{r^{\l}(\lambda)} \xrightarrow[\lambda \to 0]{} 0 \,,\,\forall\, \l=1,\dots,L \,.
\end{equation}

\begin{proof}
To simplify the discussion we set
\begin{align*}
    \x^{\l} &\coloneqq \Phi_{\hm^{\l}}(\x^{0}) \,,\, \l = 1, \dots, L \,, \\
    \x^{\l}_{\lambda} &\coloneqq \Phi_{\lambda, \hm^{\l}}(\x^{0}) \,,\, \l = 1, \dots, L \,, 
\end{align*}
to be the \textit{quantized representations} and the \textit{regularised representations} of $\x^{0}$ in the $\l$-th layer, respectively.
First, we note that
\begin{equation*}
\begin{split}
    \| \x^{\l}_{\lambda} - \x^{\l} \|
    &\coloneqq \left( \sum_{i=1}^{n_{\l}} | x^{\l}_{\lambda, i} - x^{\l}_{i} |^{2} \right)^{\frac{1}{2}} \\
    &\leq \sqrt{n_{\l}} \max_{i \in \{ 1, 2, \dots, n_{\l} \}} \{ | x^{\l}_{\lambda, i} - x^{\l}_{i} | \} \,,
\end{split}
\end{equation*}
where $x^{\l}_{\lambda, i}$ and $x^{\l}_{i}$ denote the $i$-th components of the regularised and quantized representations, respectively.
We define
\begin{equation*}
    \tilde{\imath} \coloneqq \mathop{\argmax}_{i \in \{ 1, 2, \dots, n_{\l} \}} \{ | x^{\l}_{\lambda, i} - x^{\l}_{i} | \} \,.
\end{equation*}
Therefore, since $n_{\l}$ is arbitrary but finite, a sufficient condition for \eqref{eq:pointwise_convergence_thesis} is
\begin{equation}\label{eq:pointwise_convergence_thesis_onedim}
    \frac{| x^{\l}_{\lambda, \tilde{\imath}} - x^{\l}_{\tilde{\imath}} |}{r^{\l}(\lambda)} \xrightarrow[\lambda \to 0]{} 0 \,.
\end{equation}
To simplify the notation, in the following we will omit the subscript index $\tilde{\imath}$.

First, we conveniently rewrite \eqref{eq:pointwise_convergence_thesis_onedim} according to the definition of limit:
\begin{equation}\label{eq:pointwise_convergence_thesis_onedim_limdef}
    \forall\, \eps > 0 \,,\, \,\exists\, \tilde{\lambda} > 0 \,:\, | x^{\l}_{\lambda} - x^{\l} | < \eps r^{\l}(\lambda) \,,\, \,\forall\, 0 < \lambda < \tilde{\lambda} \,.
\end{equation}
Then, we argue by induction.

To prove the base step ($\l = 1$) we need to consider two cases: $x^{1} = 0$ and $x^{1} = 1$.
First, we suppose $x^{1} = \sigma(A_{\m^{1}}(\x^{0})) = 1$; this implies that $A_{\m^{1}}(\x^{0}) < 0$.
Property \eqref{eq:regularisation_parameter_H4} implies $x^{1}_{\lambda} \geq x^{1}$, which implies $| x^{1}_{\lambda} - x^{1} | = x^{1}_{\lambda} = \sigma_{\lambda^{1}}(A_{\m^{1}}(\x^{0}))$.
Then, we can apply $\sigma_{\lambda^{1}}^{-1}$ to both sides of the inequality in \eqref{eq:pointwise_convergence_thesis_onedim_limdef}, thus obtaining the following condition:
\begin{equation*}
    \forall\, \eps > 0 \,,\, \,\exists\, \tilde{\lambda} > 0 \,:\, A_{\m^{1}}(\x^{0}) < \sigma_{\lambda^{1}}^{-1}(\eps r^{1}(\lambda)) \,,\, \,\forall\, 0 < \lambda < \tilde{\lambda} \,,
\end{equation*}
whose validity is guaranteed by hypothesis \eqref{th:pointwise_convergence_hp_i}.
Now, we analyze the case $x^{1} = 1$.
Property \eqref{eq:regularisation_parameter_H4} implies $x^{1}_{\lambda} \leq x^{1}$, hence $| x^{1}_{\lambda} - x^{1} | = 1 - x^{1}_{\lambda} = 1 - \sigma_{\lambda^{1}}(A_{\m^{1}}(\x^{0}))$.
In this case, condition \eqref{eq:pointwise_convergence_thesis_onedim_limdef} becomes
\begin{equation}\label{eq:pointwise_convergence_thesis_onedim_limdef_transf}
    \forall\, \eps > 0 \,,\, \,\exists\, \tilde{\lambda} > 0 \,:\, 1 - \sigma_{\lambda^{1}}(A_{\m^{1}}(\x^{0})) < \eps r^{1}(\lambda) \,,\, \,\forall\, 0 < \lambda < \tilde{\lambda} \,.
\end{equation}
We have two sub-cases: $A_{\m^{1}}(\x^{0}) > 0$ and $A_{\m^{1}}(\x^{0}) = 0$.
In the first sub-case, by rearranging terms and applying $\sigma_{\lambda^{1}}^{-1}$ to both sides of \eqref{eq:pointwise_convergence_thesis_onedim_limdef_transf}, we derive the condition
\begin{equation*}
    \forall\, \eps > 0 \,,\, \,\exists\, \tilde{\lambda} > 0 \,:\, \sigma_{\lambda^{1}}^{-1}(1 - \eps r^{1}(\lambda)) < A_{\m^{1}}(\x^{0}) \,,\, \,\forall\, 0 < \lambda < \tilde{\lambda} \,,
\end{equation*}
which is granted by hypothesis \eqref{th:pointwise_convergence_hp_ii}.
In the second sub-case, we can just divide both sides of \eqref{eq:pointwise_convergence_thesis_onedim_limdef_transf} by $r^{1}(\lambda)$ and obtain the condition
\begin{equation*}
    \forall\, \eps > 0 \,,\, \,\exists\, \tilde{\lambda} > 0 \,:\, \frac{1 - \sigma_{\lambda^{1}}(0)}{r^{1}(\lambda)} < \eps \,,\, \,\forall\, 0 < \lambda < \tilde{\lambda} \,,
\end{equation*}
which holds by hypothesis \eqref{th:pointwise_convergence_hp_iii}.

We now proceed to the inductive step ($\l > 1$).
We have two possibilities for $x^{\l}$:
\begin{enumerate}[label=(\Alph*)]
    \item\label{th:pointwise_convergence_caseA} $x^{\l} = 0$;
    \item\label{th:pointwise_convergence_caseB} $x^{\l} = 1$.
\end{enumerate}
We start with the case \ref{th:pointwise_convergence_caseA}.
We observe that
\begin{equation}\label{eq:pointwise_convergence_caseA}
    s^{\l}_{\lambda} - s^{\l}
    = A_{\m^{\l}}(\x^{\l-1}_{\lambda}) - A_{\m^{\l}}(\x^{\l-1}) 
    = A_{\m^{\l}}(\x^{\l-1}_{\lambda} - \x^{\l-1}) 
    \xrightarrow[\lambda \to 0]{} 0 \,,
\end{equation}
since $A_{\m^{\l}}$ is linear and $\| \x^{\l-1}_{\lambda} - \x^{\l-1} \| \xrightarrow[\lambda \to 0]{} 0$ by the inductive hypothesis.
With reference to \eqref{eq:heaviside_zero2one}, case \ref{th:pointwise_convergence_caseA} implies that $s^{\l} < 0$.
Together with \eqref{eq:pointwise_convergence_caseA}, this implies that
\begin{equation*}
    \exists\, \lambda^{*} = \lambda^{*}(s^{\l}) > 0 \,:\, s^{\l}_{\lambda} < -\frac{|s^{\l}|}{2} < 0 \,,\, \,\forall\, 0 < \lambda < \lambda^{*} \,.
\end{equation*}
Since $x^{\l} = 0$ and $x^{\l}_{\lambda} = \sigma_{\lambda^{\l}}(s^{\l}_{\lambda}) \geq 0$, condition \eqref{eq:pointwise_convergence_thesis_onedim_limdef} can be rewritten as
\begin{equation*}
    \forall\, \eps > 0 \,,\, \,\exists\, \tilde{\lambda} > 0 \,:\, \sigma_{\lambda^{\l}}(s^{\l}_{\lambda}) < \eps r^{\l}(\lambda) \,,\, \,\forall\, 0 < \lambda < \tilde{\lambda} \,.
\end{equation*}
Due to the monotonicity of $\sigma_{\lambda^{\l}}$, we have $\sigma_{\lambda^{\l}}(s^{\l}_{\lambda}) < \sigma_{\lambda^{\l}}(-|s^{\l}|/2) \,,\, \,\forall\, 0 < \lambda < \lambda^{*}$.
Therefore, a sufficient condition to guarantee the convergence is that
\begin{equation*}
    \forall\, \eps > 0 \,,\, \,\exists\, 0 < \tilde{\lambda} \leq \lambda^{*} \,:\, -\frac{|s^{\l}|}{2} < \sigma_{\lambda^{\l}}^{-1}(\eps r^{\l}(\lambda)) \,,\, \,\forall\, 0 < \lambda < \tilde{\lambda} \,.
\end{equation*}
This is granted for every $s^{\l} < 0$ by \eqref{th:pointwise_convergence_hp_i}.

We now move to the case \ref{th:pointwise_convergence_caseB}.
This case ($x^{\l} = 1$) might originate from two sub-cases:
\begin{enumerate}[label=(\roman*)]
    \item\label{th:pointwise_convergence_caseBi} $s^{\l} > 0$;
    \item\label{th:pointwise_convergence_caseBii} $s^{\l} = 0$.
\end{enumerate}

The proof of subcase \ref{th:pointwise_convergence_caseBi} is similar to the proof for the case \ref{th:pointwise_convergence_caseA}.
Given $s^{\l} > 0$, since $s^{\l}_{\lambda} \xrightarrow[\lambda \to 0]{} s^{\l}$ by the inductive hypothesis, we have that
\begin{equation*}
    \exists\, \lambda^{*} = \lambda^{*}(s^{\l}) > 0 \,:\, 0 < \frac{s^{\l}}{2} < s^{\l}_{\lambda} \,.
\end{equation*}
Then, since $x^{\l} = 1$ and $x^{\l} = \sigma_{\lambda^{\l}}(s^{\l}_{\lambda}) \leq 1$, we can rewrite \eqref{eq:pointwise_convergence_thesis_onedim_limdef} as
\begin{equation*}
    \forall\, \eps > 0 \,,\, \,\exists\, \tilde{\lambda} > 0 \,:\, 1 - \sigma_{\lambda^{\l}}(s^{\l}_{\lambda}) < \eps r^{\l}(\lambda) \,,\, \,\forall\, 0 < \lambda < \tilde{\lambda} \,.
\end{equation*}
Since $\sigma_{\lambda^{\l}}(s^{\l}_{\lambda}) > \sigma_{\lambda^{\l}}(s^{\l}/2)$, a sufficient condition to get convergence is that
\begin{equation*}
    \forall\, \eps > 0 \,,\, \,\exists\, 0 < \tilde{\lambda} < \lambda^{*} \,:\, \frac{s^{\l}}{2} > \sigma_{\lambda^{\l}}^{-1}(1 - \eps r^{\l}(\lambda)) \,,\, \,\forall\, 0 < \lambda < \tilde{\lambda} \,.
\end{equation*}
This is guaranteed for every $s^{\l} > 0$ by \eqref{th:pointwise_convergence_hp_i}.

The case \ref{th:pointwise_convergence_caseBii} is more delicate, since $s^{\l}_{\lambda}$ can be positioned in two ways with respect to $s^{\l} = 0$: \begin{enumerate}[label=(\alph*)]
    \item\label{th:pointwise_convergence_caseBiia} $\lambda > 0$ is such that $s^{\l}_{\lambda} \geq 0$;
    \item\label{th:pointwise_convergence_caseBiib} $\lambda > 0$ is such that $s^{\l}_{\lambda} < 0$.
\end{enumerate}

In case \ref{th:pointwise_convergence_caseBiia}, it is sufficient to note that the monotonicity of $\sigma_{\lambda^{\l}}$ implies $1 - \sigma_{\lambda^{\l}}(s^{\l}_{\lambda}) \leq 1 - \sigma_{\lambda^{\l}}(0)$, since $s^{\l}_{\lambda} \geq 0$.
Then, condition \eqref{eq:pointwise_convergence_thesis_onedim_limdef} can be rewritten as
\begin{equation*}
    \forall\, \eps > 0 \,,\, \,\exists\, \tilde{\lambda} > 0 \,:\, \frac{1 - \sigma_{\lambda^{\l}}(0)}{r^{\l}(\lambda)} < \eps \,,\, \,\forall\, 0 < \lambda < \tilde{\lambda} \,.
\end{equation*}
This is guaranteed by \eqref{th:pointwise_convergence_hp_iii}.

To prove the last case \ref{th:pointwise_convergence_caseBiib}, we first observe that
\begin{equation*}
\begin{split}
    s^{\l}_{\lambda}
    &= s^{\l}_{\lambda} - s^{\l} \\
    &= \left( \langle \x^{\l-1}_{\lambda}, \w^{\l} \rangle + b^{\l} \right) - \left( \langle \x^{\l-1}, \w^{\l} \rangle + b^{\l} \right) \\
    &= \langle \x^{\l-1}_{\lambda} - \x^{\l-1}, \w^{\l} \rangle
\end{split}
\end{equation*}
(which follows from $s^{\l} = 0$) and apply the Cauchy-Schwartz inequality to obtain the following upper bound:
\begin{equation}\label{eq:pointwise_convergence_cauchyschwartz}
    |s^{\l}_{\lambda}| \leq \| \x^{\l-1}_{\lambda} - \x^{\l-1} \| \, \| \w^{\l} \| \,.
\end{equation}
Then, we rewrite \eqref{eq:pointwise_convergence_thesis_onedim_limdef} as
\begin{equation*}
    \forall\, \eps > 0 \,,\, \,\exists\, \tilde{\lambda} > 0 \,:\, -s^{\l}_{\lambda} < -\sigma_{\lambda^{\l}}^{-1}(1 - \eps r^{\l}(\lambda)) \,,\, \,\forall\, 0 < \lambda < \tilde{\lambda} \,.
\end{equation*}
Observation \eqref{eq:pointwise_convergence_cauchyschwartz} allows us to write a slightly stronger but sufficient condition for convergence:
\begin{equation*}
    \forall\, \eps > 0 \,,\, \,\exists\, \tilde{\lambda} > 0 \,:\, \| \x^{\l-1}_{\lambda} - \x^{\l-1} \| \, \| \w^{\l} \| \leq -\sigma_{\lambda^{\l}}^{-1}(1 - \eps r^{\l}(\lambda)) \,,\, \,\forall\, 0 < \lambda < \tilde{\lambda} \,,
\end{equation*}
where we used the fact that $-s^{\l}_{\lambda} = |s^{\l}_{\lambda}|$ (since $s^{\l}_{\lambda} < 0$).
The inner inequality can be rewritten as
\begin{equation}\label{eq:pointwise_convergence_caseBiib_transf}
    \frac{\| \x^{\l-1}_{\lambda} - \x^{\l-1} \|}{-\sigma_{\lambda^{\l}}^{-1}(1 - \eps r^{\l}(\lambda))} \leq \frac{1}{\| \w^{\l}\|} \,;
\end{equation}
since $\w^{\l}$ is fixed but arbitrary (it is part of the parameter $\m^{\l}$), the term on the right can be arbitrarily small, and therefore a sufficient condition to ensure \eqref{eq:pointwise_convergence_caseBiib_transf} for $\lambda$ small enough is
\begin{equation}\label{eq:pointwise_convergence_caseBiib_suffcond}
    \forall\, \eps > 0 \,,\, \frac{\| \x^{\l-1}_{\lambda} - \x^{\l-1} \|}{-\sigma_{\lambda^{\l}}^{-1}(1 - \eps r^{\l}(\lambda))} \xrightarrow[\lambda \to 0]{} 0 \,.
\end{equation}
By the inductive hypothesis (where we set $\eps = 1$),
\begin{equation*}
    \exists\, \tilde{\lambda}^{\l-1} > 0 \,:\, \| \x^{\l-1}_{\lambda} - \x^{\l-1} \| < r^{\l-1}(\lambda) \,.
\end{equation*}
Therefore, \eqref{th:pointwise_convergence_hp_iv} enforces the convergence of the upper bound in the following inequality:
\begin{equation*}
    \frac{\| \x^{\l-1}_{\lambda} - \x^{\l-1} \|}{-\sigma_{\lambda^{\l}}^{-1}(1 - \eps r^{\l}(\lambda))} \leq \frac{r^{\l-1}(\lambda)}{-\sigma_{\lambda^{\l}}^{-1}(1 - \eps r^{\l}(\lambda))} \,,
\end{equation*}
and therefore \eqref{eq:pointwise_convergence_caseBiib_suffcond} follows.
This completes the proof of the theorem.
\end{proof}

\begin{figure}[ht]
    \centering
    \includegraphics[width=0.6\textwidth]{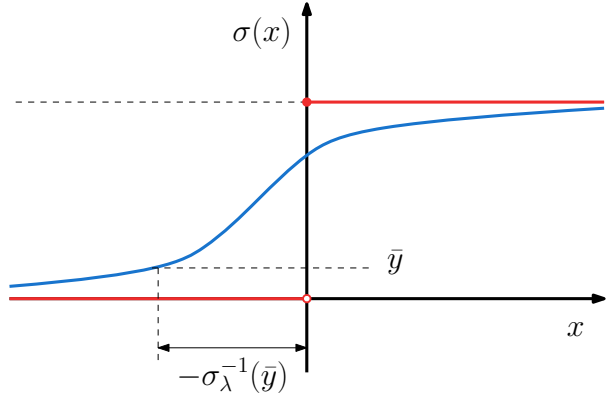}
    \caption{The Heaviside function $H^{+}$ (red) is approximated by a regularised function $\sigma_{\lambda}$ (blue); the geometric meaning of $\sigma^{-1}_{\lambda}$ is illustrated.}\label{fig:regularised_activation_function}
\end{figure}
\end{thm}

We conclude this section by providing two examples of regularisation of Heaviside functions.
In the first example, we consider a one-parameter family of sigmoid activations and show that it satisfies the rate convergence property for suitably chosen convergence rates.
In the second, we consider a simple network composed of two layers with Heaviside activation functions, together with a regularised counterpart which does not satisfy all the assumptions \eqref{th:pointwise_convergence_hp_i}-\eqref{th:pointwise_convergence_hp_iv} of Theorem~\ref{th:pointwise_convergence}, and show that the pointwise convergence of the regularised networks to the initial network fails on a large subset of the input space.

\begin{xmp}\label{xmp:pointwise_convergence_xmp}
Let $\lambda > 0$ denote a real regularisation parameter.
For each $\l = 1, \dots, L$, define $\lambda^{\l} \coloneqq \lambda^{\l}(\lambda) > 0$ such that $\lambda^{\l} \to 0$ as $\lambda \to 0$, and consider the regularised functions
\begin{equation*}
\begin{split}
    \sigma_{\lambda^{\l}} \,:\,
    \R &\to \R \\
    s &\to \frac{1}{1 + e^{-\frac{s + \lambda^{\l}}{(\lambda^{\l})^2}}} \,.
\end{split}
\end{equation*}
We have
\begin{equation*}
    \sigma_{\lambda^{\l}}^{-1}(x) = -(\lambda^{\l})^{2} \log\left( \frac{1 - x}{x} \right) - \lambda^{\l} \,,\, x \in (0, 1) \,.
\end{equation*}

Fix $\eps > 0$, and rewrite \eqref{th:pointwise_convergence_hp_i} and \eqref{th:pointwise_convergence_hp_ii} as
\begin{align*}
    \sigma_{\lambda^{\l}}^{-1}(\eps r^{\l}(\lambda))
    &= - (\lambda^{\l})^{2} \log\left( \frac{1 - \eps r^{\l}(\lambda)}{\eps r^{\l}(\lambda)} \right) - \lambda^{\l} \\
    &= - (\lambda^{\l}) \log(1 - \eps r^{\l}(\lambda)) + (\lambda^{\l})^{2} \log(\eps) + (\lambda^{\l})^{2} \log(r^{\l}(\lambda)) - \lambda^{\l} \,, \\
    \sigma_{\lambda^{\l}}^{-1}(1 - \eps r^{\l}(\lambda))
    &= - (\lambda^{\l})^{2} \log\left( \frac{\eps r^{\l}(\lambda)}{1 - \eps r^{\l}(\lambda)} \right) - \lambda^{\l} \\
    &= - (\lambda^{\l})^{2} \log(\eps) - (\lambda^{\l})^{2} \log(r^{\l}(\lambda)) + (\lambda^{\l})^{2} \log(1 - \eps r^{\l}(\lambda)) - \lambda^{\l} \,.
\end{align*}
Since $(\lambda^{\l})^{2} \log(\eps)$, $(\lambda^{\l})^{2} \log(1 - \eps r^{\l}(\lambda))$ and $\lambda^{\l}$ tend to zero when $\lambda \to 0$, the only requirement for convergence is
\begin{equation*}
    (\lambda^{\l})^{2} \log(r^{\l}(\lambda)) \xrightarrow[\lambda \to 0]{} 0 \,.
\end{equation*}
By rewriting \eqref{th:pointwise_convergence_hp_iii} as
\begin{equation*}
\begin{split}
    \frac{1 - \sigma_{\lambda^{\l}}(0)}{r^{\l}(\lambda)}
    &= \frac{1 - \left( 1 + e^{-\frac{1}{\lambda^{\l}}} \right)}{\left( 1 + e^{-\frac{1}{\lambda^{\l}}} \right) r^{\l}(\lambda)} \\
    &= - \frac{1}{\left( 1 + e^{-\frac{1}{\lambda^{\l}}} \right) \left( e^{\frac{1}{\lambda^{\l}}} \right) r^{\l}(\lambda)}
\end{split}
\end{equation*}
one sees that the condition
\begin{equation*}
    e^{\frac{1}{\lambda^{\l}}} r^{\l}(\lambda) \xrightarrow[\lambda \to 0]{} \infty
\end{equation*}
(since $1 + e^{-\frac{1}{\lambda^{\l}}} \to 1$ as $\lambda \to 0$). On the other hand, by analysing assumption \eqref{th:pointwise_convergence_hp_iv} one has that
\begin{align*}
    \frac{r^{\l-1}(\lambda)}{\sigma_{\lambda^{\l}}^{-1}(1 - \eps r^{\l}(\lambda))} &= \frac{r^{\l-1}(\lambda)}{- (\lambda^{\l})^{2} \log(1 - \eps r^{\l}(\lambda)) + (\lambda^{\l})^{2} \log(\eps) + (\lambda^{\l})^{2} \log(r^{\l}(\lambda)) - \lambda^{\l}}\\
    &\sim \frac{r^{\l-1}(\lambda)}{ (\lambda^{\l})^{2} \log(r^{\l}(\lambda)) - \lambda^{\l}}\,,
\end{align*}
hence if we assume that $r^\l(\lambda)$ goes to zero like a power of $\lambda^\l$, we conclude that the required infinitesimality condition is implied by
\begin{equation*}
    \frac{r^{\l-1}(\lambda)}{\lambda^{\l}}\to 0\,.  
\end{equation*}
Therefore, a possible choice is to set $r^{L}(\lambda) = \lambda^{L} = \lambda$ and $r^{\l}(\lambda) = \lambda^{\l} = \lambda^{2 (L - \l)}$, for $\l = 1,\dots, L-1$.
This choice describes an \textit{annealing schedule} where convergence is faster in the first layers (i.e., for small $\l$) and progressively slower in successive layers (i.e., for $\l$ large). It is interesting to notice that this type of annealing has already been experimentally found, see \cite{Severa2019, Spallanzani2019}.
\end{xmp}

\begin{xmp}\label{xmp:pointwise_convergence_counterxmp}
Let $\lambda$ be a positive real parameter.
Let $\sigma = H^+$ be the Heaviside function defined in \eqref{eq:heaviside_zero2one}, and define the following piece-wise affine approximation
\begin{equation}\label{eq:pointwise_convergence_counterxmp_regularisedact}
\begin{split}
    \sigma_{\lambda} \,:\,
    \R &\to [0, 1] \\
    s &\mapsto
    \begin{cases}
        \lambda \,, &\text{if } s \leq -\lambda^{2} \,, \\
        \frac{(1 - \lambda)}{\lambda^{2}} s + 1 \,, &\text{if } -\lambda^{2} < s \leq 0 \,, \\
        1 \,, &\text{if } 1 < s \,.
    \end{cases}
\end{split}
\end{equation}
Although in theory $\sigma_{\lambda}$ should be differentiable and strictly increasing, we have made this simplified choice to better illustrate the convergence issues.

Clearly, $\sigma_{\lambda}$ is non-decreasing and bounded for every $\lambda$, and converges to \eqref{eq:heaviside_zero2one} when $\lambda$ goes to zero.
Now, we consider the following QNN:
\begin{equation*}
    \Phi(x^{0}) \coloneqq \sigma\left( \sigma(x^{0} w^{1} + b^{1}) w^{2} + b^{2} \right) \,.
\end{equation*}
In particular, we suppose $w^{1} \neq 0$ and set $w^{2} = -1$ and $b^{2} = 0$ to get
\begin{equation}\label{eq:pointwise_convergence_counterxmp_qnn}
    \Phi(x^{0}) = \sigma\left( -\sigma(x^{0} w^{1} + b^{1}) \right) \,.
\end{equation}
The function $\Phi$ is the characteristic function of the set
\begin{equation*}
    S_{1} \coloneqq \{ x^{0} \,|\, x^{0} w^{1} + b^{1} < 0 \}\,.
\end{equation*}
Consider now the regularised version of \eqref{eq:pointwise_convergence_counterxmp_qnn} obtained using \eqref{eq:pointwise_convergence_counterxmp_regularisedact}:
\begin{equation*}
    \Phi_{\lambda, \hm^{2}}(x^{0}) \coloneqq \sigma_{\lambda}\left( -\sigma_{\lambda}(x^{0} w^{1} + b^{1}) \right) \,.
\end{equation*}
For every $x^{0} \in S_{1}$, define
\begin{equation*}
    \tilde{\lambda}(x^{0}) \coloneqq \sup \{ \lambda \,|\, \sigma_{\lambda}(x^{0} w^{1} + b^{1}) = \lambda \} \,;
\end{equation*}
it follows that
\begin{equation*}
    \sigma_{\lambda}(x^{0} w^{1} + b^{1}) = \lambda \,,\, \,\forall\, 0 < \lambda < \tilde{\lambda}(x^{0}) \,.
\end{equation*}
Therefore, we have
\begin{equation*}
\Phi_{\lambda, \hm^{2}}(x^{0})
    = \sigma_{\lambda}(-\lambda) 
    = \lambda
\end{equation*}
for every $\lambda$ such that $0 < \lambda < \min \{ 1, \tilde{\lambda}(x^{0}) \}$, since $-\lambda < -\lambda^{2}$ when $\lambda < 1$. 
But then we have that
\begin{equation*}
\lim_{\lambda \to 0} \Phi_{\lambda, \hm^{2}}(x^{0})
    = \lim_{\lambda \to 0} \lambda 
    = 0,\quad \text{while}\ \  
    \Phi(x^{0})=1, \qquad \forall\, x^0 \in S_1\,.
\end{equation*}
This shows that the pointwise convergence fails on the whole half-line $S_1$.
\begin{figure}
    \centering
    \includegraphics[width=\textwidth]{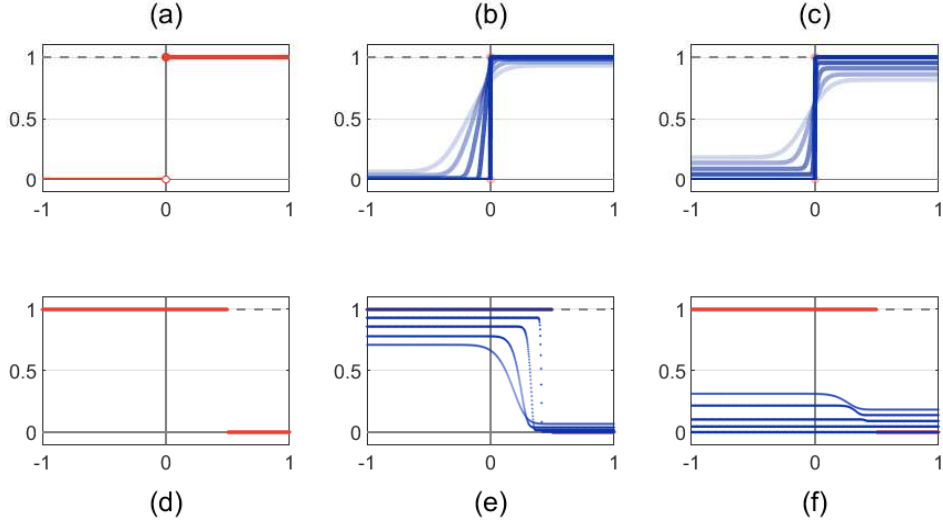}
    \caption{Top row: the Heaviside function $H$ (a) is approximated point-wise by a regularised function $\sigma_{\lambda}$ that satisfies the rate-convergence hypothesis (b) and by a function $\tilde{\sigma}_{\lambda}$ that does not satisfy it (c).
        Bottom row: the composition $\Phi(x) = H(-H(2x - 1))$ (d) is approximated by $\Phi_{\lambda}(x) = \sigma_{\lambda}(-\sigma_{\lambda}(2x - 1))$ (e) and by $\tilde{\Phi}_{\lambda}(x) = \tilde{\sigma}_{\lambda}(-\tilde{sigma}_{\lambda}(2x - 1))$; whereas $\Phi_{\lambda}$ converges point-wise to $\Phi$, $\tilde{\Phi}_{\lambda}$ fails to converge on the whole half-line $\{ x \in \R \,|\, x < 1/2 \}$.}\label{fig:pointwise_convergence}
\end{figure}
\end{xmp}

\subsection{Quantizing the weights}
The training of a quantized neural network requires a (local) optimisation of a loss function defined on a discrete set of parameters.
If considered in its full generality, such a problem is of combinatorial type, thus practically intractable when the number of parameters is very high.
However, one might embed the discrete parameter set into a continuous space where the computation of gradients as well as the definition of a gradient descent algorithm become more natural.
At the same time, one should take into account that the choice of the embedding might strongly influence the behaviour of an optimisation algorithm.
Even more, one should be warned against the possible problems that may occur when a quantized minimiser has to be determined from a continuous minimiser.
The following example shows that a na\"{i}ve projection of regularised (optimal) parameters onto quantized counterparts can lead to arbitrarily large errors.

\begin{xmp}\label{xmp:continuous_to_quantized_parameters}
Let $D > 2$ be a real parameter, $N \geq 2$ be an integer, and consider the following data set sampled from $\R^{2} \times \{ 0, 1 \}$:
\begin{equation*}
\mathcal{D} \coloneqq \Big\{(\x^{(i)},y^{(i)}):\ i=1,\dots,2N\Big\}
\end{equation*}
with $\x^{(i)} = (x^{(i)}_1,x^{(i)}_2)$ and $y^{(i)}$ defined as
\begin{align*}
&x^{(1)}_1 = -D,\quad x^{(1)}_2 = 2,\quad y^{(1)} = 1;\\ 
&x^{(i)}_1 = D,\quad x^{(i)}_2 = -2^{2-i}, \quad y^{(i)} = 1\qquad \text{for }i=2\dots,N;\\ 
&x^{(N+1)}_1 = D,\quad x^{(N+1)}_2 = -2,\quad y^{(N+1)} = 0;\\ 
&x^{(N+i)}_1 = -D,\quad x^{(N+i)}_2 = 2^{N+2-i}, \quad y^{(N+i)} = 0\qquad \text{for }i=2\dots,N.
\end{align*}
This data set is linearly separable, as one can infer from Figure~\ref{fig:quantized_weights_data_set}.

Our goal is to find parameters $\w = (w_{1}, w_{2})'$ such that the artificial neuron implementing
\begin{equation}\label{eq:quantized_weights_neuron}
\begin{split}
    f_{\w} \,:\,
    \R^{2} &\to \{ 0, 1 \} \\
    \x &\mapsto H(x_{1} w_{1} + x_{2} w_{2})
\end{split}
\end{equation}
classifies correctly as many data points as possible.
We constrain $b = 0$ to simplify the problem.
We measure the accuracy of a solution as the fraction of points which are wrongly classified by \eqref{eq:quantized_weights_neuron}:
\begin{equation*}
    \Loss(f_{\w}) \coloneqq \frac{\# \big\{ (\x^{(i)}, y^{(i)}) \in \mathcal{D} \,|\, f_{\w}(\x^{(i)}) \neq y^{(i)} \big\}}{2N} \,.
\end{equation*}

If we constrain $w_{i} \in \{ -1, 0, 1 \} \,,\, i = 1, 2$ (i.e., we are looking for a neuron using ternary weights), then the problem is a discrete optimisation.
One might be tempted to solve it by relaxing the constraint and allowing the parameters to take values in $[-1, 1]$; for example, by computing the expected values of discrete random variables taking values in $\{ -1, 0, 1 \}$.

Solutions with optimal accuracy ($\Loss(f_{\w}) = 0$) can be found taking $\w$ in the angle comprised between the half-lines
\begin{equation}\label{eq:quantized_weights_half-line_low}
\begin{cases}
    x_{2} = -\frac{x^{(N+1)}_{1} - x^{(1)}_{1}}{x^{(N+1)}_{2} - x^{(1)}_{2}} x_{1} \\
    x_{1} > 0 \,,
\end{cases}
\end{equation}
and
\begin{equation}\label{eq:quantized_weights_half-line_high}
\begin{cases}
    x_{2} = -\frac{x^{(2)}_{1} - x^{(N+2)}_{1}}{x^{(2)}_{2} - x^{(N+2)}_{2}} x_{1} \\
    x_{1} > 0 \,.
\end{cases}
\end{equation}
The first line corresponds to the direction of the normal to the line touching both $\x^{(1)}$ and $\x^{(N+1)}$; the second line corresponds to the direction of the normal to the line touching both $\x^{(N+2)}$ and $\x^{(2)}$.
These conditions imply the following constraints on the components of $\w$:
\begin{equation*}
    w_{1} > 0\,,\qquad
    w_{2} > 0 \,,\qquad
    \frac{D}{2} < \frac{w_{2}}{w_{1}} < D \,.
\end{equation*}
An example is given in Figure~\ref{fig:quantized_weights_regularised_optimal}.

But when such solutions are projected onto their quantized counterparts, the performance of the resulting classifier might deteriorate.
For example, when $D > 4$, the \textit{regularised} vector $\w$ must belong to the triangle comprised between the half-lines \eqref{eq:quantized_weights_half-line_low}, \eqref{eq:quantized_weights_half-line_high} and the line $x_{2} = 1$.
Remembering that the projections of each component must take values in $\{ -1, 0, 1 \}$, these constraints are such that each optimal solution is projected on $\w = (0, 1)$.
An example is given in Figure~\ref{fig:quantized_weights_regularised_projected}: every point in $\mathcal{D}$ except for $\x^{(1)}$ and $\x^{(N+1)}$ are assigned an incorrect label, yielding
\begin{equation*}
    \Loss(f_{\w})
     = 1 - \frac{1}{N} \xrightarrow[N \to \infty]{} 1 \,.
\end{equation*}

An example optimal quantized solution is instead depicted in Figure~\ref{fig:quantized_weights_optimal}, where $\w = (1, 0)$.
The corresponding classifier has almost perfect performance:
\begin{equation*}
    \Loss(f_{\w})
    = \frac{1}{N} \xrightarrow[N \to \infty]{} 0 \,.
\end{equation*}

\begin{figure}[ht]
\centering
\begin{subfigure}{.4\textwidth}
    \centering
    \caption{}\label{fig:quantized_weights_data_set}
    \includegraphics[width=.8\linewidth]{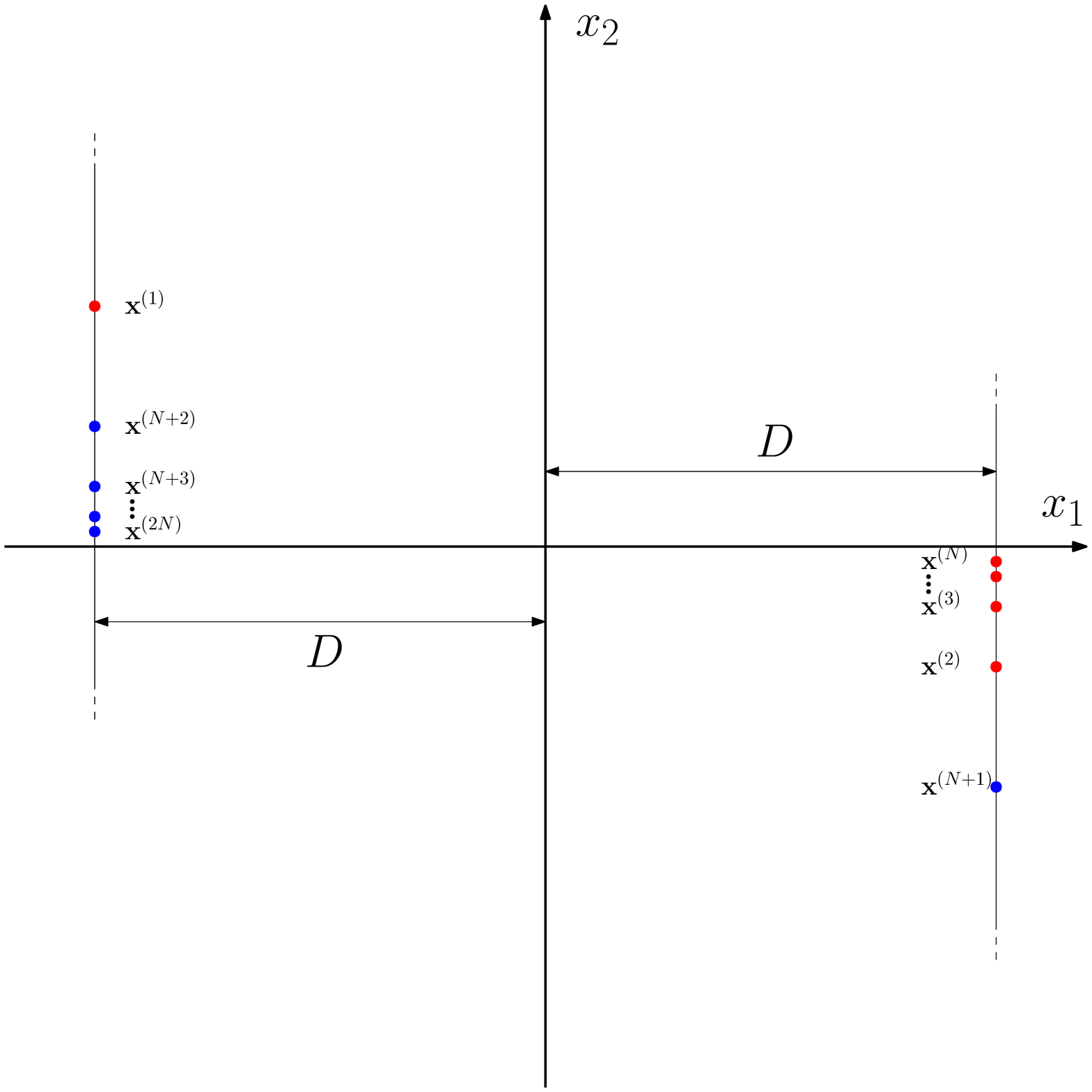}
\end{subfigure}%
\begin{subfigure}{.4\textwidth}
    \centering
    \caption{}\label{fig:quantized_weights_optimal}
    \includegraphics[width=.8\linewidth]{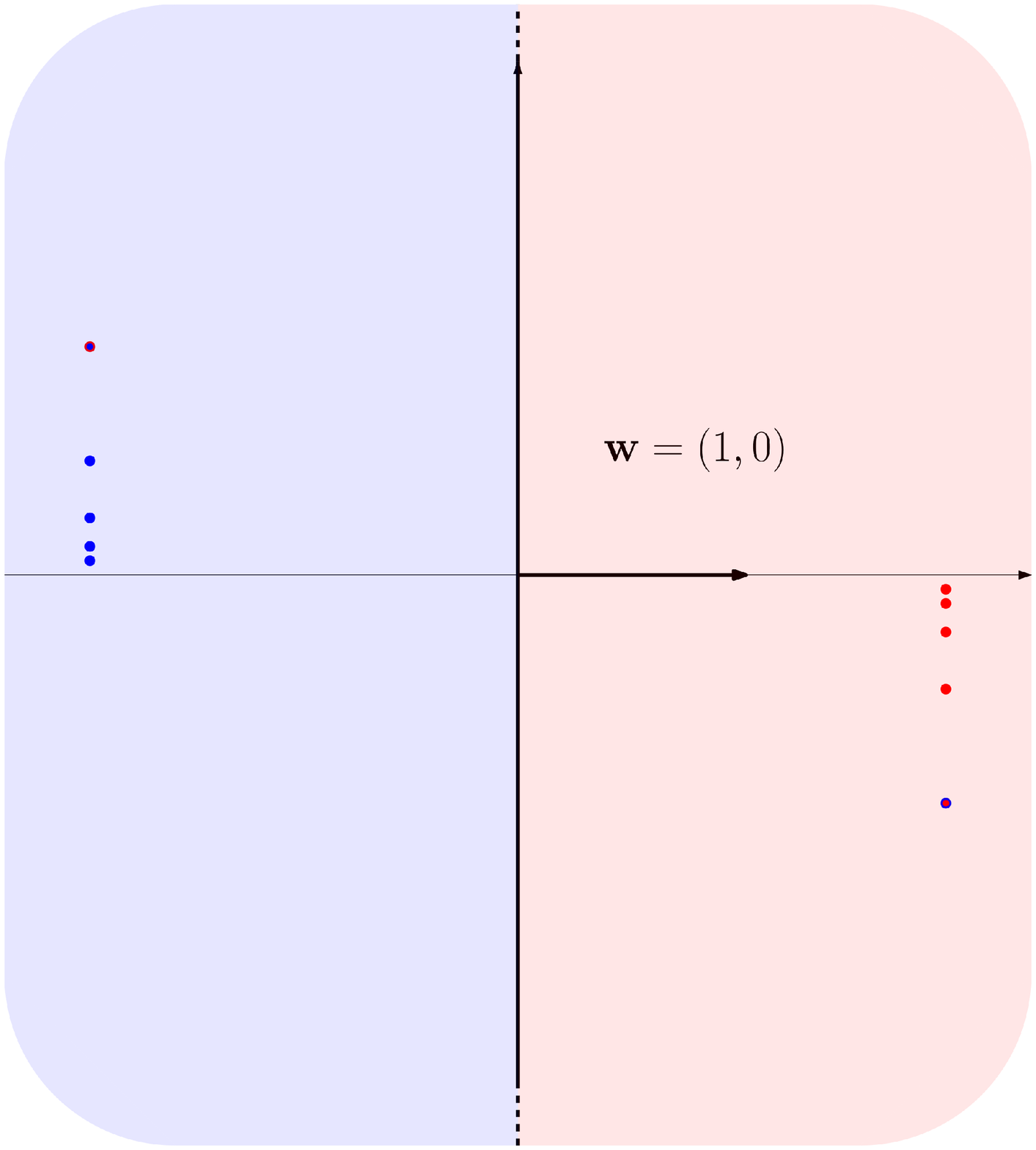}
\end{subfigure}
\begin{subfigure}{.4\textwidth}
    \centering
    \includegraphics[width=.8\linewidth]{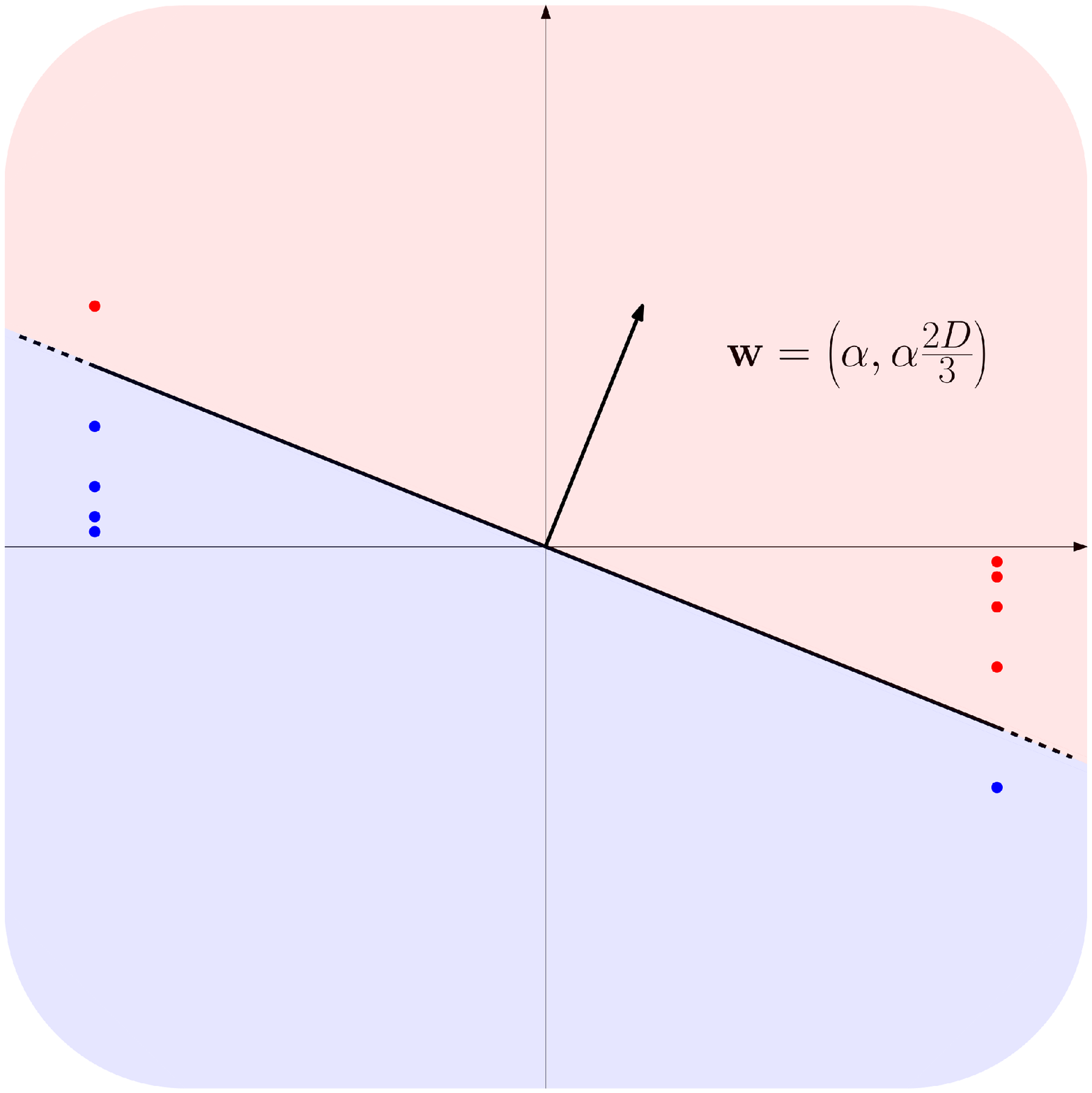}
    \caption{}\label{fig:quantized_weights_regularised_optimal}
\end{subfigure}%
\begin{subfigure}{.4\textwidth}
    \centering
    \includegraphics[width=.8\linewidth]{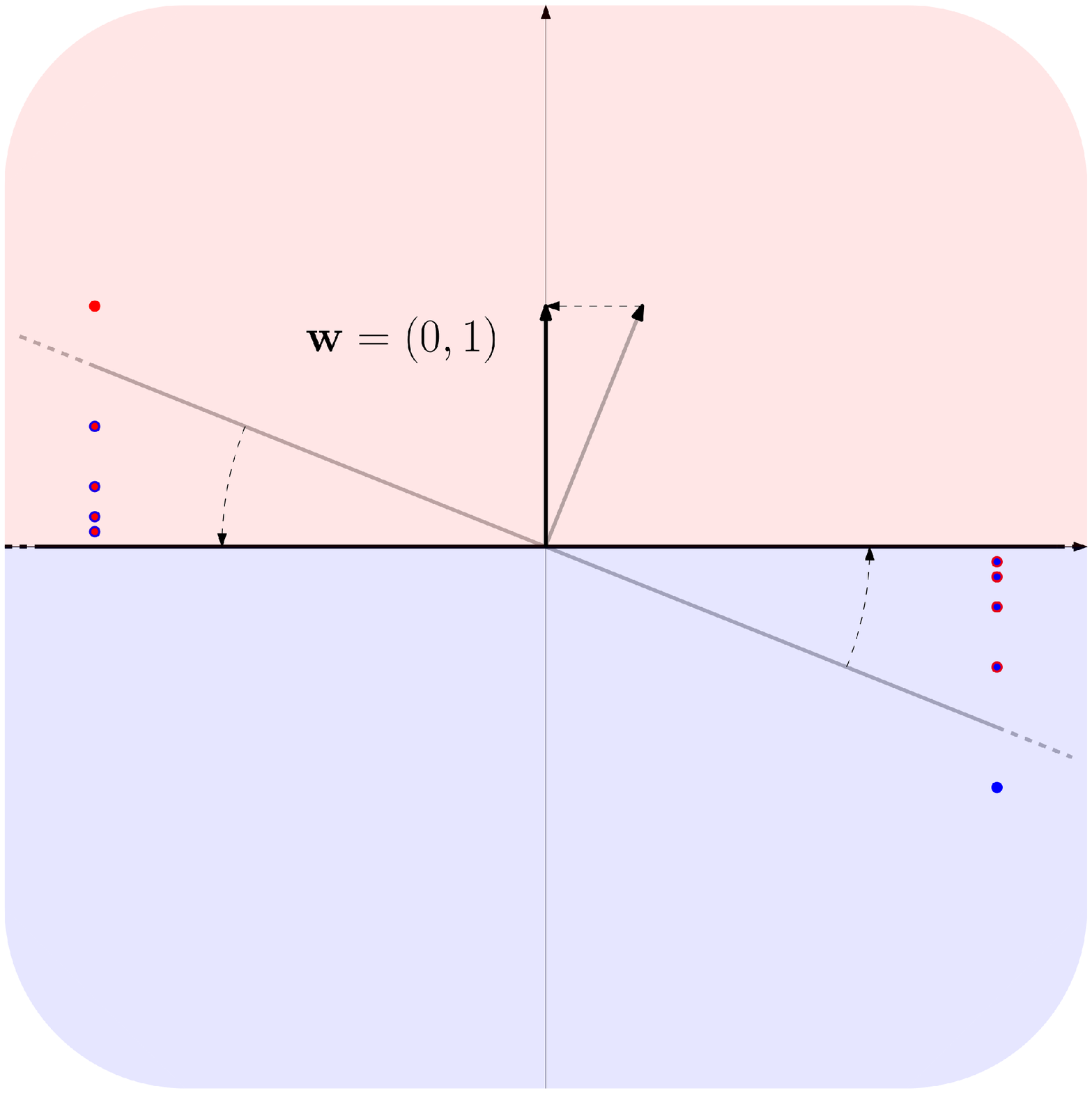}
    \caption{}\label{fig:quantized_weights_regularised_projected}
\end{subfigure}
\caption{}\label{fig:quantized_weights}
\end{figure}
\end{xmp}

\section{Conclusions}
\label{sec:conclusions}

Hardware-oriented optimisations are fundamental for accelerating the execution of deep neural networks and for deploying them to embedded systems or MCUs.
However, they pose several problems, especially for what concerns the learning process via standard stochastic gradient descent, as various difficulties related to the lack of differentiability of the networks arise in a very natural way.
The two typical processes where these difficulties originate from are:
\begin{itemize}
    \item the simplification of the activation functions, leading to point-wise non-differentiability (as in ReLUs) or even to the presence of jump discontinuities coupled with almost-everywhere zero derivative (as in Heaviside functions);
    \item the parameter quantization, which produces a strong constraint and limitation on gradient-based training.
\end{itemize}

The present study enlightens various possibilities as well as limitations in the application of the above optimisation paradigm.
Its main contributions are hereafter summarised.
\begin{enumerate}
    \item Theorem~\ref{th:expressivity} answers in the affirmative the question of whether QNNs are as expressive as general DNNs.
        Specifically, it shows that every $\Lambda$-Lipschitz function defined on a hypercube can be uniformly approximated by a three-layer QNN with an explicit bound on the number of parameters that are needed to define the network.
        Even though the complexity of the space $\Lip_{\Lambda}$ necessarily makes our bound suffer from the curse of dimensionality, we believe that our result represents a starting point for a deeper analysis of QNN-related function spaces satisfying better approximation properties. 
    \item Theorem~\ref{th:uniform_compositional_approximation} proposes a layer-wise stochastic regularisation of a generic continuous (but possibly non-differentiable) feedforward neural network and provides a quantitative $L^{\infty}$ estimate that, in the special case of Lipschitz-continuous networks, reduces to a weighted sum of $L^{1}$ variances of the layers' parameters, with coefficients given by suitable products of the Lipschitz constants of the original layer maps.
        This estimate suggests that noise variances should be decreased according to a suitable \textit{annealing schedule}.
        Moreover, it allows the design of a principled training procedure where locally optimal parameters are obtained via stochastic gradient descent applied to the regularised network.
    \item In Section~\ref{sec:discontinuity}, we have considered the problem of approximating and training a discontinuous neural network.
        Theorem~\ref{th:pointwise_convergence} guarantees that under a suitable set of assumptions on the approximating sequence of smooth activations (the \textit{rate-convergence} property), the smoothed network approximates the original discontinuous network in a pointwise sense (i.e., fixing the input $\x^{0}$ and the layers' parameters $\m^{1}, \dots, \m^{L}$).
        The geometric idea behind the technical assumptions \eqref{th:pointwise_convergence_hp_i}-\eqref{th:pointwise_convergence_hp_iv} is that the smoothed Heaviside activation functions must not only converge to the Heaviside function $H^+$ in a pointwise sense but also that the corresponding graphs should converge faster to the horizontal parts than to the vertical part of the (multi-)graph of $H^+$.
        This fact is crucial, as illustrated in Example~\ref{xmp:pointwise_convergence_xmp} and Example~\ref{xmp:pointwise_convergence_counterxmp}.
        On the other hand, as Example~\ref{xmp:continuous_to_quantized_parameters} shows, one cannot expect that a na\"{i}ve projection of continuous parameters onto quantized counterparts can behave well in terms of QNNs optimisation.
\end{enumerate}

\end{document}